\documentclass[default,iicol]{sn-jnl}

\jyear{2021}%

\theoremstyle{thmstyleone}%
\theoremstyle{thmstyletwo}%

\theoremstyle{thmstylethree}%

\usepackage{epsfig}
\usepackage{graphicx}
\usepackage{amsmath}
\usepackage{amssymb}
\usepackage{multirow}
\usepackage{makecell}
\usepackage{subfigure}
\usepackage{array}
\newcommand{\tabincell}[2]{\begin{tabular}{@{}#1@{}}#2\end{tabular}} 
\usepackage{footnote}

\begin{document}

\title[Article Title]{Learning Continuous Face Representation with Explicit Functions}



\author[1,2,3,4]{\fnm{Liping} \sur{Zhang}}\email{zliping@semi.com}
\equalcont{These authors contributed equally to this work.}
\author*[1,2,3,4]{\fnm{Weijun} \sur{Li}}\email{wjli@semi.com}
\equalcont{These authors contributed equally to this work.}

\author[1,2,3,4]{\fnm{Linjun} \sur{Sun}}\email{sunlinjun@semi.com}

\author[1,2,3]{\fnm{Lina} \sur{Yu}}\email{yulina@semi.com}

\author[1,2,3,4]{\fnm{Xin} \sur{Ning}}\email{ningxin@semi.com}

\author[1,2,3,4]{\fnm{Xiaoli} \sur{Dong}}\email{dongxiaoli@semi.com}

\author[1,2,3]{\fnm{Jian} \sur{Xu}}\email{xujian@semi.com}

\author[1,2,3]{\fnm{Hong} \sur{Qin}}\email{qinh@semi.com}

\affil[1]{\orgdiv{Institute of Semiconductors}, \orgname{Chinese Academy of Sciences}, \orgaddress{\city{Beijing}, \postcode{100083}, \country{China}}}

\affil[2]{\orgdiv{Center of Materials Science and Optoelectronics Engineering \& School of Microelectronics}, \orgname{University of Chinese Academy of Sciences}, \orgaddress{ \city{Beijing}, \postcode{100049}, \country{China}}}

\affil[3]{\orgdiv{Beijing Key Laboratory of Semiconductor Neural Network Intelligent Sensing and Computing Technology}, \orgname{Institute of Semiconductors}, \orgaddress{\city{Beijing}, \postcode{100083}, \country{China}}}

\affil[4]{\orgdiv{Cognitive Computing Technology Joint Laboratory}, \orgname{Wave Group}, \orgaddress{\city{Beijing}, \postcode{100083}, \country{China}}}


\abstract{ How to represent a face pattern? While it is presented in a continuous way in our visual system, computers often store and process the face image in a discrete manner with 2D arrays of pixels. In this study, we attempt to learn a continuous representation for face images with explicit functions. First, we propose an explicit model (EmFace) for human face representation in the form of a finite sum of mathematical terms, where each term is an analytic function element. Further, to estimate the unknown parameters of EmFace, a novel neural network, EmNet, is designed with an encoder-decoder structure and trained using the backpropagation algorithm, where the encoder is defined by a deep convolutional neural network and the decoder is an explicit mathematical expression of EmFace. Experimental results show that EmFace has a higher representation performance on faces with various expressions, postures, and other factors, compared to that of other methods. Furthermore, EmFace achieves reasonable performance on several face image processing tasks, including face image restoration, denoising, and transformation.}




\maketitle

\section{Introduction}\label{sec1}

Human faces have highly personalized and rich pattern characteristics that not only serve as an identity but also convey different attributes, such as gender, emotions, age, intentions, attractiveness, and ethnic background \cite{oruc2019face}. Many studies have been conducted based on face images, including face recognition, face detection, facial landmark detection, facial expression recognition, face generation, etc. \cite{ guo2019survey, zhang2016joint, zhang2018joint, karras2019style}. The achievements have been widely used in intelligent security inspection, human–computer interaction, identity recognition, social platforms, face beautification, and many other applications. A fundamental research area is exploring appropriate ways to express, describe, and analyze face images.

\begin{figure}[t]
	\centering
	\includegraphics[width=1.0\linewidth]{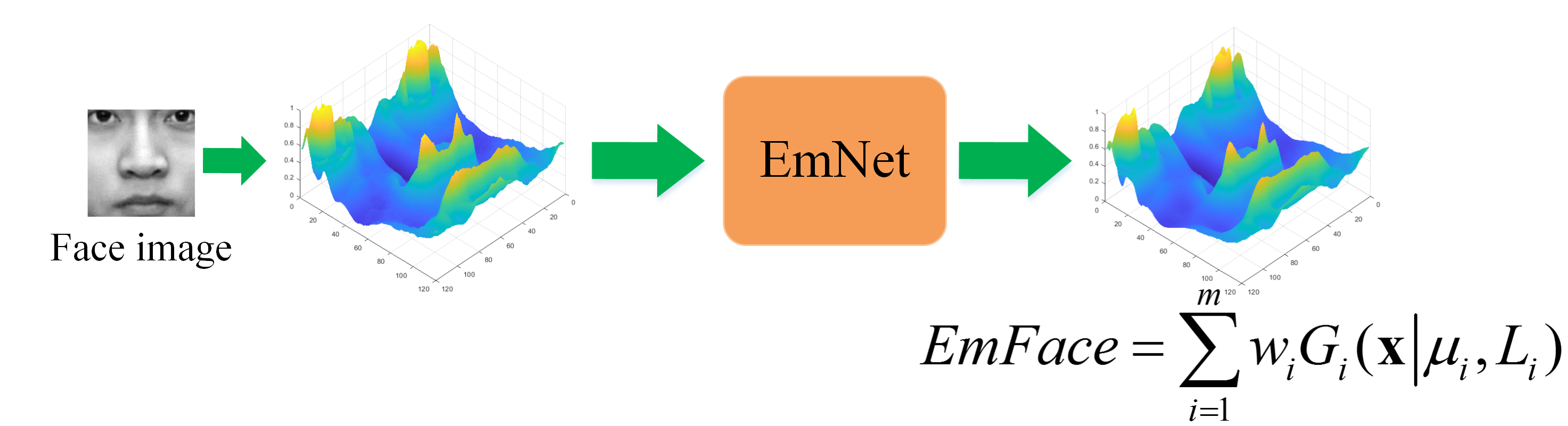}
	\caption{The illustration of our method, which Takes a gray face image as a 2D surface, then introduces an explicit model (EmFace) with a fixed number of function elements to model the surface. EmNet is the parameters solving network of EmFace.}\label{fig:MultiGaussForFace}
\end{figure}

A review of the research on face representation shows that existing approaches are commonly based on feature analysis or compression reconstruction of face images, which represent face pattern in a discrete manner with 2D arrays of pixels. A variety of face analysis methods have been proposed in recent years \cite{ ahonen2004face, shan2004review, lindeberg2012scale, shu2011histogram, turk1991eigenfaces, belhumeur1997eigenfaces, duan2018context, lu2017simultaneous, masi2018deep, sun2014deep, sun2015deeply, taigman2014deepface, schroff2015facenet}, which can be broadly categorized into traditional shallow and deep neural network representations.

Traditional shallow representations extract low-dimensional representations, multilevel and high-dimensional extensions or encoding codebook to describe face images. However, these approaches attempt to represent face images through single- or double-layer features (filter responses, histograms, or binary codes), which inevitably limits their robustness against the complex nonlinear facial appearance variations in a variety of real applications. Deep neural network (DNN) methods, such as convolutional neural networks (CNNs), can learn multiple levels of image features that correspond to different levels of abstraction, though a cascade of multiple layers of convolution processing units. Using large labeled datasets, researchers adopt the end-to-end training mode to map image data to representation forms, which could be suitable for particular tasks. Although DNN representations have reshaped the landscape of facial image research \cite{ masi2018deep, sun2014deep, sun2015deeply, taigman2014deepface, schroff2015facenet}, these models lack transparency and interpretability, that is, they cannot explain exactly what they learned, why they emulated a certain task, or why they made a particular decision.

Mathematical modeling is the use of mathematical logic methods and language to build scientific or engineering models, wherein a complex research object is simplified as a mathematical representation.  For several centuries, mathematical modeling has been established as a valuable means for scientists to understand the world and discover laws \cite{ schmidt2009distilling, iten2020discovering, raissi2019physics, lake2017building, 9157823}. As the most familiar body part, the face shares common morphological features of the eyes, nose, and mouth. However, the face is also an aggregation with a complex structure in which all facial features are personally distinct and can be assembled differently. Although the topological structure of each face is similar, individual identities can still be distinguished by the facial features. Furthermore, owing to age, environment, psychology, and other factors, even face images of the same person can exhibit a rich variety of appearances at different times, including a smile, sadness, anger, disgust, surprise, and fear \cite{2018A}. Owing to such differences and dynamics, a face image can be regarded as a multivariable nonlinear complex system. Moreover, there is no prior knowledge of either the expressions or parameters. Therefore, mathematical representation of the human face remains an open challenge.

In this context, the goal of this study is to construct a general explicit model that models human face images, and to learn the parameters of this explicit model from data to transform a face image from a discrete pixel space into a continuous function space (Fig. \ref{fig:MultiGaussForFace}). Hence, we consider a face image as a two-dimensional (2D) surface, model it as an analytic function, and approximate that function by a linear combination of function elements; we term this formulation as an explicit model for face image (EmFace). EmFace itself is defined by the quantity and formula of function elements, and the parameters of EmFace are a collection of the parameters of each element and their weights.

Various methods can be applied to solve the parameters of mathematical model, such as genetic programming \cite{koza1992genetic}, evolutionary computation \cite{back2018evolutionary}, and Expectation-Maximization (EM) algorithm \cite{sridharan2014gaussian}. In our previous study \cite{GMFace}, a neural network called GmNet was designed using function elements as neurons, and the backpropagation algorithm was used to effectively solve the coefficients. However, the algorithm required a number of iterations to solve the model parameters for each particular face, which was time-consuming and non-universal. To address the drawbacks of GmNet, we propose a network EmNet, which considers function parameters as the feature values of discrete pixel values in a continuous function space; therefore, parameter solving can be achieved using a special autoencoder. It can not only provide the mathematical model EmFace for the human face but also complete the parameter solution for all input face images. Thus, it is a universal face-modeling network. To the best of our knowledge, no such study has been attempted thus far. In addition, we verify the feasibility of the proposed method.

As an encoder-based method, EmNet is entirely self-supervised. Only a collection of function elements and their desired number (${N}$, usually 80) are required. The output face model is concise (${6N}$ values) and enables the transformation of a face image using a series of parameter calculations.

The main contributions of this study are as follows:

\begin{itemize}
	\item [1)]
	In contrast to analyzing face images in discrete pixel spaces, we model face images in a continuous function space, and propose an explicit model for face image representation (EmFace) in the form of the quantity and formula of the function elements.
	
	\item [2)]
	We design a neural network called EmNet for the parameter solving of EmFace. In addition, EmNet takes the parameters of EmFace as the feature values of the input image, which provides a reference for other mathematical modeling tasks.	
	
	\item [3)]
	We demonstrate the potential of our method for several applications, including face image modeling, restoration, denoising, and transformation.
\end{itemize}

The remainder of this paper is organized as follows. Related work on face representations, mathematical modeling based on neural networks and image coding representations is presented in Section 2. Section 3 provides a detailed description of the proposed method. The experiments are presented in Section 4. Conclusions with a summary and outlook for future work are presented in Section 5.

\section{Related Work}\label{sec2}

A brief introduction to three related topics is presented in this section: 1) face representation, 2) mathematical modeling based on neural network, and 3) image coding representation.

\subsection{Face representation}

Many studies have been conducted on 2D face image analysis aimed at extracting feature representations to resolve particular tasks, such as face recognition, expression synthesis, and facial animation. These face representation includes traditional shallow and deep neural network (DNN) representations.


Existing traditional shallow representations can be mainly classified into three categories: holistic approaches, hand-crafted local descriptors, and learning-based local approaches. Holistic approaches \cite{turk1991eigenfaces, belhumeur1997eigenfaces, 0Nonlinear} learn feature mapping to retain the statistical information of face images. Handcrafted local descriptors \cite{ahonen2004face, shan2004review, lindeberg2012scale, shu2011histogram} describe the structure pattern of each local patch through the invariant properties of local filtering. Learning-based local approaches \cite{lu2015learning, duan2018context} also focus on local patterns. Their basic idea is to learn local filters or encode codebooks for higher distinctiveness and compactness. For DNN representations, based on large labeled datasets, researchers have used an end-to-end learning mode to map image data to a feature set, which could be suitable for particular tasks \cite{masi2018deep}. Resently, more mainstream network architectures, such as the DeepID series \cite{sun2014deep, DeepID2, sun2015deeply}, FaceNet \cite{taigman2014deepface}, SphereFace \cite{liu2017sphereface}, and ArcFace \cite{deng2019arcface} have been designed to obtain face representation that can be used for discrimination purposes. 

The aforementioned methods have been successfully applied to a variety of tasks related to human face images. However, neither traditional shallow representations nor DNN representations can provide a simplified representation of the human face itself. In contrast to these methods, we do not aim to extract feature representations to resolve specific tasks. Instead, we attempt to model a face image into a continuous function space and approximate the mathematical model of the human face.

\subsection{Mathematical modeling based on neural network}

Recently, mathematical modeling based on neural networks has garnered increasing interest. It can be used to determine natural laws and then model them mathematically using massive data. Iten  et al. \cite{iten2020discovering} proposed a neural network architecture named SciNet to extract physical concepts, such as conservation laws of angular momentum, quantum state degree of freedom, and heliocentric model of the solar system. In addition, researchers have attempted to model existing scientific knowledge using DNN technology, such as Hamilton’s law \cite{schmidt2009distilling}, Lagrange’s theorem \cite{schmidt2009distilling}, and Korteweg-de Vries equation \cite{raissi2019physics}. In \cite{2021Encoder}, a novel neural network Encoder-X was developed for both explicit and implicit polynomial fittings. In three-dimensional (3D) shape representation, deep implicit functions \cite{9008389, 9157823} have been shown to be highly effective. They represent an input observation based on directly learning the continuous implicit function and can achieve state-of-the-art results for several 3D shape reconstruction tasks. A network was proposed in \cite{9008389} to encode shapes into implicit functions, which are represented as a mixture of scaled axis-aligned anisotropic 3D Gaussians. In these frontier studies, DNN technology plays a key role, which enables researchers to discover scientific laws from experimental data.

\subsection{Image coding representation}

A number of image coding methods have proven useful in image representation, although they are not specific to face image representation. Because there are no published reports on explicit models for face image modeling, we compared most of our results with those of image coding methods such as principal component analysis (PCA) \cite{turk1991eigenfaces}, sparse coding (SC) \cite{Wright2009Robust, Bao2016Dictionary}, and autoencoder-based methods \cite{0Reducing}.

Turk and Pentland proposed a method for face representation known as the classical eigenface using PCA in 1991 \cite{turk1991eigenfaces}. All 2D face images are represented as the projection coefficients of the image on the first ${n}$ eigenvectors. SC is a classical signal processing technique and is commonly used as an efficient technique for image reconstruction in the computer vision field \cite{2013Laplacian}. SC aims to construct succinct representations of the input image.

The autoencoder (AE) is a type of approach that automatically learns features in an unsupervised manner. It is commonly used for data dimensionality reduction and denoising \cite{0Reducing} by minimizing the reconstruction loss. To process image data, a stacked convolutional autoencoder (CAE) is pre-trained in a layer-wise unsupervised procedure to initialize a CNN \cite{2014Stacked}. Pascal Vincent et al. introduced a new training principle, called denoising autoencoder (DAE), to learn representations robust to partial corruption of the input \cite{2008Extracting, 2010Stacked}. A sparse autoencoder (SAE) \cite{ng2011sparse} can learn better features by imposing a sparsity constraint on the hidden units. A variational automatic encoder (VAE) \cite{kingma2013auto} is a data generation model that can produce data similar to training samples. Although AE-based methods can effectively encode data as a vector of hidden variables, the hidden variables are meaningless and cannot provide an explicit mathematical representation for face images.

\begin{figure}[b]
	\centering
	\includegraphics[width=1.0\linewidth]{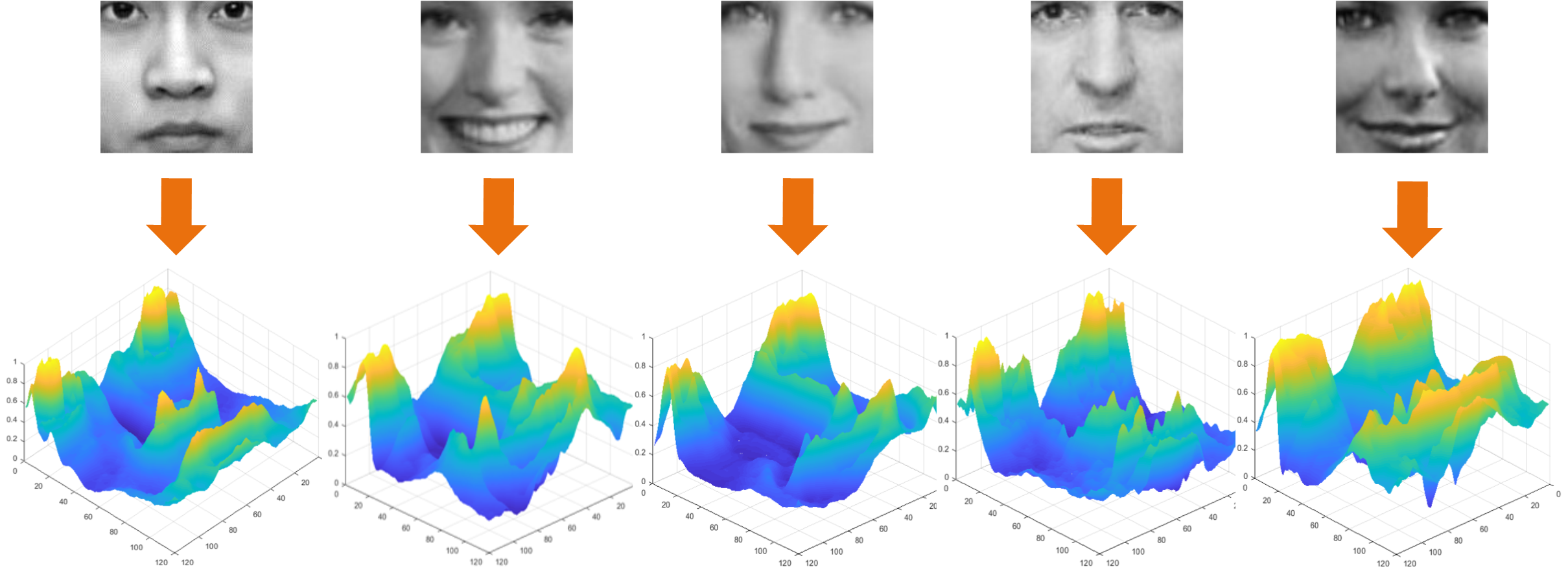}
	\caption{2D surfaces for some face image samples.}\label{fig:2D_surface}
\end{figure}

\section{Explicit face image representation}\label{sec3}

\subsection{Face pattern analysis}


\begin{figure*}[t]
	\centering
	\includegraphics[width=1.0\linewidth]{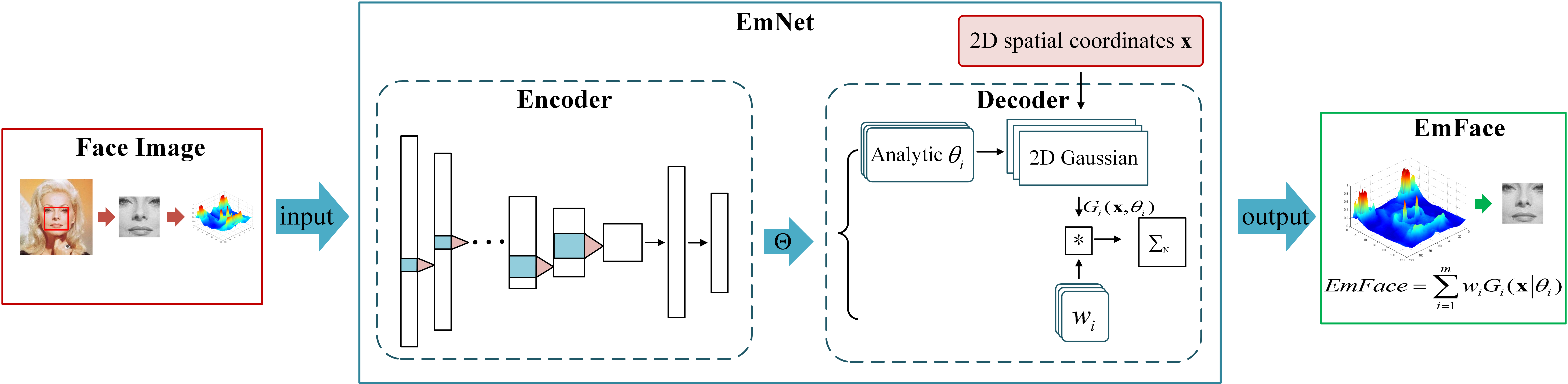}
	\caption{Overview of our proposed EmNet for face image modeling. The input to our system is the surface of a cropped face. It starts with an encoder based on a CNN structure, and the output of the CNN is a vector with fixed dimensionality. Subsequently, this vector is interpreted as a set of function elements with parameters that define an explicit function. An MSE loss enforces the EmFace to reconstruct the input image as much as possible in the training phase. }\label{fig:EF-Facenet framework}		
\end{figure*}

Usually, in the analysis of 2D face images, only the pixel intensity is utilized \cite{zhao2018pixel, shen2017novel, asthana2014pixels}, whereas the spatial coordinates and relation between pixel intensity and spatial coordinates are often ignored. Herein, we propose an alternative solution for face image representation, which models analytical expressions to obtain deep insights into face patterns. To achieve this goal, we consider 2D spatial coordinates as independent variables and pixel intensity as a dependent variable based on coordinates; subsequently, the gray face image can be expressed as a 2D surface, as shown in Fig. \ref{fig:2D_surface}. As illustrated, faces have a unique pattern, that is, a continuously changing surface. In particular, there is a more obvious peak-valley distribution at the eyes, nose, and mouth, whereas the cheek is gentler and smoother. It is expected that this pattern can be expressed in a function space.

\subsection{EmFace}

We aim to represent this surface as an analytic function ${F(\bf{x},\Theta )}$, where ${\bf{x}}$ is a 2D position and ${\Theta}$ is the vector of parameters. In the explicit formulation, ${F}$ is the weighted summation of a fixed number of function elements, labeled ${i \in \left[ N \right]}$, where ${N}$ is the number, and weight coefficient ${w_i}$ denotes the contribution of the ${i^{th}}$ elements to the output. Each element is a function ${f_i}$ defined by the parameter vector ${\theta _i}$.

\begin{equation}
	F({\bf{x}},\Theta ) = \sum\limits_{i \in [N]} {{w_i}{f_i}({\bf{x}},{\theta _i})}.
\end{equation}

The specific version of the function element we adopt is \emph{2D Gaussian}. Here, ${\theta _i}$ consists of a geometric center ${{\bf{\mu }} = \left[ {\begin{array}{*{20}{c}}
			{{\mu _1}}\\
			{{\mu _2}}
	\end{array}} \right]}$ and precision matrix ${{\bf{A}} = \left[ {\begin{array}{*{20}{c}}
			{{a_{11}}}&{{a_{12}}}\\
			{{a_{21}}}&{{a_{22}}}
	\end{array}} \right]}$, which is the inverse of the covariance matrix. A \emph{2D Gaussian} is mathematically expressed as follows:
\begin{equation}
	f({\bf{x}}, \theta _i) = \exp \{  - {({\bf{x}} - {\bf{\mu }})^T}{\bf{A}}({\bf{x}} - {\bf{\mu }})\}.
\end{equation}

Thus, the EmFace can be constructed for each pixel as
\begin{equation}
	EmFace({\bf{x}},{w_i},{{\bf{\mu}} _i},{{\bf{A}}_i}) = \sum\limits_{i \in [N]} {{w_i}{f_i}({\bf{x}}, {\mu _i},{{\bf{A}}_i})}.
\end{equation}

According to the definition of a \emph{2D Gaussian}, ${\bf{A}}$ in Eq. (2) is a positive-definite symmetric matrix. To guarantee the positive definiteness of ${\bf{A}}$, the elements of ${\bf{A}}$ must satisfy the following constraints:
\begin{equation}
	\left\{ {\begin{array}{*{20}{c}}
			{{a_{21}} = {a_{12}}}\\
			{{a_{11}} > 0}\\
			{{a_{22}} > 0}\\
			{{a_{11}} \times {a_{22}} - {a_{21}} \times {a_{12}} > 0}.
	\end{array}} \right.
\end{equation}

We set an arbitrary vector ${\bf{t}}{\rm{ = }}\left[ {\begin{array}{*{20}{c}}{{t_1}}\\{{t_2}}\end{array}} \right]$ and ${\left\| \rho  \right\| < 1}$ to construct matrix ${\bf{A}}$ that satisfies the positive definiteness.
\begin{equation}
	\begin{array}{l}
		{\bf{A}}{\rm{ = }}{\bf{t}}*{{\bf{t}}^T}\left\{ {\rho \left[ {\begin{array}{*{20}{c}}
					1&1\\
					1&1
			\end{array}} \right] + (1 - \rho )\left[ {\begin{array}{*{20}{c}}
					1&0\\
					0&1
			\end{array}} \right]} \right\}\\
		{\rm{  \ \  }} = \left[ {\begin{array}{*{20}{c}}
				{t_1^2}&{\rho {t_1}{t_2}}\\
				{\rho {t_1}{t_2}}&{t_2^2}
		\end{array}} \right].
	\end{array}
\end{equation}

\section{EmNet: EmFace Learning}

To solve EmFace parameters, we propose a framework, called EmNet (Fig. \ref{fig:EF-Facenet framework}), to train a DNN architecture to fit EmFace from the data. The goal of the network is to find the function parameters that best fit a 2D face image, where the loss penalizes the discrete pixel value error of each position. We cropped and normalized the face image to provide input to the network. The entire framework consists of an encoder and a decoder. As the encoder, the DNN architecture encodes the input face image into hidden variables (the vector of parameters ${\Theta}$ in our method). Next, EmFace, as the decoder, provides the mathematical model of the human face. 

\subsection{Architecture}
\begin{figure}[t]
	\centering
	\includegraphics[width=0.5\linewidth]{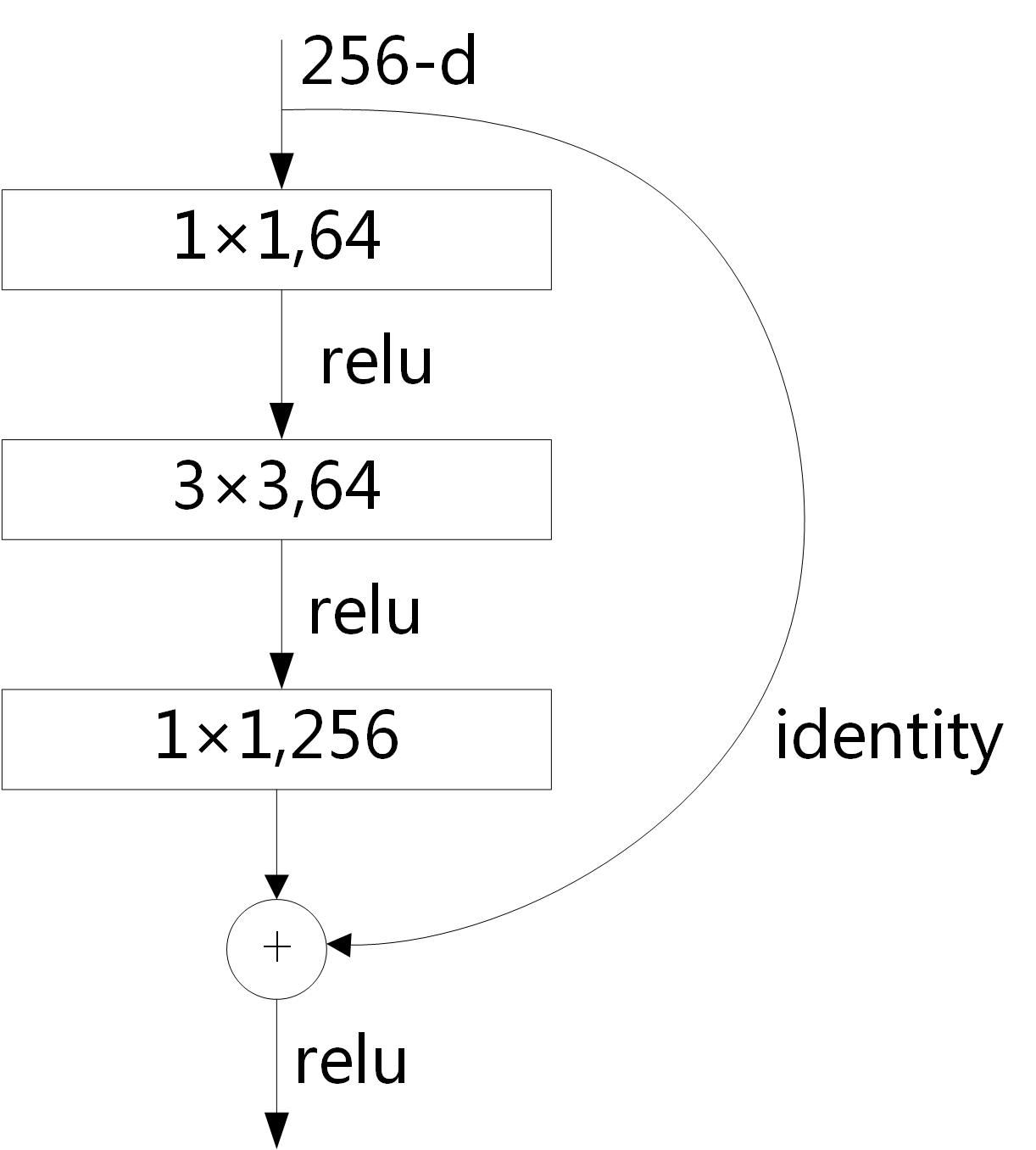}
	\caption{Three-layer bottleneck block for ResNet50.}\label{fig:Residual_block}	
\end{figure}

To learn the parameters of EmFace, we first need to encode the input face image. There are various CNN architectures for encoding images into feature vectors, such as LeNet \cite{lecun1998gradient}, AlexNet \cite{krizhevsky2012imagenet}, VGG-Net \cite{simonyan2014very}, GoogleNet \cite{szegedy2015going}, and ResNet \cite{he2016deep}. These architectures can effectively extract image features by local connections, shared weights, and down-sampling processing. We selected the ResNet50 architecture \cite{he2016deep} as the encoder in EmNet, which contains 16 three-layer bottleneck blocks, as shown in Fig. \ref{fig:Residual_block}, and one fully connected layer.

The last fully connected layer of the encoder maps the output of the convolution layer to the parameter vector ${\Theta}$, which in our experiments is (${6 \times N}$)-D. Let ${\Theta  = \left[ {{\theta _1};{\theta _2}; \cdots ;{\theta _{6N}}} \right]}$, then the parameters for ${N}$ groups of \emph{2D Gaussian} components are decoded as follows:
\begin{equation}
	{{\bf{\mu}} _i}{\rm{ = sigmoid}}\left( {\left[ {\begin{array}{*{20}{c}}
				{{\theta _{6(i-1)+1}}}\\
				{{\theta _{6(i-1)+2}}}
		\end{array}} \right]} \right) \in [0,1],
\end{equation}
\begin{equation}
	{\bf{t}_i}{\rm{ = }}\left[ {\begin{array}{*{20}{c}}
			{{\theta _{6(i-1)+3}}}\\
			{{\theta _{6(i-1)+4}}}
	\end{array}} \right],
\end{equation}
\begin{equation}
	\rho_i {\rm{ = tanh(}}{\theta _{6(i-1)+5}}{\rm{)}} \in [ - 1,1],
\end{equation}
\begin{equation}
	w_i{\rm{ = tanh(}}{\theta _{6(i-1)+6}}{\rm{)}} \in [ - 1,1].
\end{equation}

With the parameters, ${N}$ groups of \emph{2D Gaussian} components are constructed. Then set query 2D coordinates ${x}$ at arbitrary resolution, according Eqs. (3) and (5), the mathematical expression for a face image can be written explicitly.

Although EmNet is an encoder/decoder style architecture, there is no decoding stage with heavy parameters: the decoder is simply the explicit mathematical expression of EmFace. In the common encoder/decoder architecture, another DNN is used to decode the hidden variables to reconstruct the image. If a ResNet50 symmetric network is used as the decoder, the parameters of the entire network will be doubled (${25.5 \times {10^6}}$ additional parameters need to be learned). It would take a considerably longer time to train, and more importantly, would not be able to provide a mathematical model of a face image that can be written.	

\subsection{EmNet training}

EmNet for face image representation is the process of transforming the face image information from the pixel space to the parameter space of EmFace. Fig. \ref{fig:EF-Facenet framework} illustrates the process of EmFace parameter solving. To solve the parameters, the 2D surface of the face image is considered as the target, and EmNet is trained to learn the parameters. Accordingly, EmFace, given in Eq. (3), can be rewritten as
\begin{equation}
	EmFace({x_1},{x_2}) = \sum\limits_{i = 1}^m {{w_i}{f_i}({x_1},{x_2}, {{\bf{\mu }}_i},{{\bf{A}}_i})},
\end{equation}

\noindent where ${x_1} = {\frac{r}{H}}$ and ${x_2} = {\frac{c}{W}}$ are the row and column coordinates of the image after normalization, respectively, ${W}$ and ${H}$ are the width and height of the face image, and ${r\in[1,H]}$ and ${c\in[1,W]}$ are the row and column indices, respectively.

To measure the reconstruction error between EmFace and the input image, we employ two individual loss functions; one is used to measure the global error, and the other is used for the local error:
\begin{equation}
	{L_2}{\rm{ = }}\sum\limits_{{x_1} = \frac{1}{H}}^1 {\sum\limits_{{x_2} = \frac{1}{W}}^1 {(EmFace({x_1},{x_2}) - I({x_1},{x_2}))} },
\end{equation}
\begin{equation}
{L_\infty } = \mathop {\max }\limits_{{x_1} \in \left[ {\frac{1}{H},1} \right],{x_2} \in \left[ {\frac{1}{W},1} \right]} \left\| {EmFace({x_1},{x_2}) - I({x_1},{x_2})} \right\|
\end{equation}

\noindent where ${L_2}$ denotes the mean square error (MSE), ${L_\infty}$ is the peak absolute error (PAE), ${W}$ and ${H}$ are the width and height of the face image, respectively. To balance the two loss functions, a scaling factor ${\alpha}$ is set. Thus, the global error measured by MSE is considered, as well as the local error, which is measured using the PAE. The total loss function used for optimizing the parameters of EmNet is constructed as follows:
\begin{equation}
	L{\rm{oss = }}{L_2}{\rm{ + }}\alpha {L_\infty }.
\end{equation}

After computing the loss value of EmNet, the back-propagation algorithm \cite{rumelhart1986learning} is used to update the parameter ${\beta}$ of the encoder until the network converges.
\begin{equation}
	\beta {\rm{ = }}\beta {\rm{ - }}\eta \frac{{\partial Loss}}{{\partial \beta }},
\end{equation}

\noindent where ${\eta}$ is the learning rate. In particular, we used adaptive moment estimation (Adam) \cite{kingma2014adam} as the optimization algorithm in our EmNet training.

\section{Applications}\label{sec4}

The goal of our method is learning continuous explicit functions to represent the face patterns. In this section, we introduce the method for applying our EmFace on several face image processing tasks, including face restoration, denoising, and transformation. Some applications, like face recognition, need to obtain features that can be used for discrimination purposes. So how to design methods to use our EmFace against such a identification task is the future research direction.

\subsection{Face image restoration and denoising}

EmNet first codes a face image into a parameter vector; this process can be regarded as the feature extraction. Subsequently, the feature values are used as the parameters of the explicit model for the face image. The face image here refers to a normal face image captured in wild conditions with large intra-class variations, such as varying poses, expressions, illuminations, and backgrounds. To enable EmNet to model a complete face pattern in case of partial corruption of the input data, we train it to reconstruct a clean ``repaired'' input face image from a corrupted, partially destroyed one. After the network converges, we can achieve facial image restoration and denoising using EmNet. This is motivated by the design of DAE: a good representation is expected to capture stable structures from a partial observation only \cite{2010Stacked}. An example of this is the ability of the human visual system to recognize partially occluded or corrupted images.

\begin{figure}[t]
	\centering
	\includegraphics[width=1.0\linewidth]{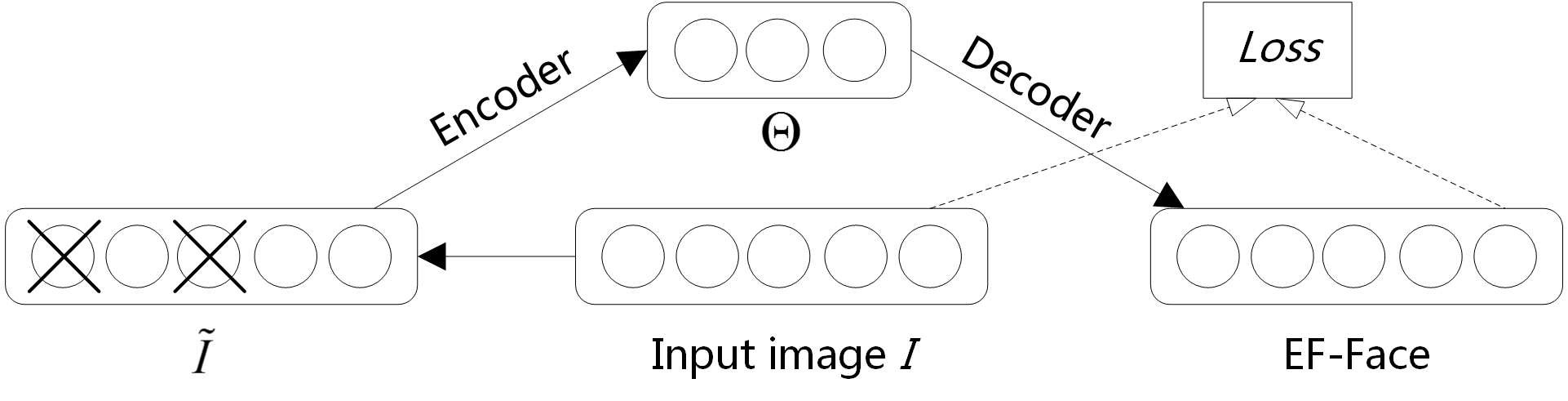}
	\caption{Training an EmNet for face image restoration and denoising. Input face image ${I}$ is stochastically corrupted (the circle represents the pixel value) to ${\tilde I}$. Subsequently, EmNet maps it to parameter ${\Theta}$ and attempts to reconstruct ${I}$ using EmFace. The reconstruction error between EmFace and ${I}$ was used to train EmNet. }\label{fig:denoising}	
\end{figure}

The implementation details are presented in Fig. \ref{fig:denoising}. The input face image ${I}$ is first corrupted to obtain a partially destroyed version ${\tilde I}$ by means of stochastic mapping ${\tilde I \sim {q_{\cal D}}\left( {\tilde I\left\| I \right.} \right)}$. The corrupted input ${\tilde I}$ is then mapped by the encoder, which is a ResNet50 architecture, to the vector of parameters ${\Theta}$. For each pixel, we construct EmFace from ${\Theta}$ using Eq. (3). As earlier, the encoder is trained to minimize the loss, as defined in Eq. (13), i.e., to have EmFace as close as possible to the uncorrupted input face image ${I}$. However, the key difference is that EmFace can reconstruct a face pattern from a corrupted input face image.

\subsection{Face Image Transformation}

EmFace can model a face image into a continuous function space. In this continuous function space, face image transformations, such as translation, scaling, and rotation, become considerably simpler. The interpolation operation that traverses all pixels can be replaced by the mathematical calculation of the function parameters. The following is a derivation of the transformation operation using EmFace.

\emph{1) Image Translation}

With the translational vector ${{\bf{\bar x}} = \left[ {\begin{array}{*{20}{c}} {{{\bar x}_1}}\\
			{{{\bar x}_2}}
	\end{array}} \right]}$, the image translation can be implemented by adjusting each geometric center ${{{\bf{\mu }}_i}}$ as follows:
\begin{equation}
	\begin{array}{c}
		EmFace({\bf{x}} - {\bf{\bar x}})\\
		= \sum\limits_{i = 1}^m {{w_i}\exp \{  - {{({\bf{x}} - {\bf{\bar x}} - {{\bf{\mu }}_i})}^T}{{\bf{A}}_i}({\bf{x}} - {\bf{\bar x}} - {{\bf{\mu }}_i})\} } \\
		= \sum\limits_{i = 1}^m {{w_i}\exp \{  - {{({\bf{x}} - ({{\bf{\mu }}_i} + {\bf{\bar x}}))}^T}{{\bf{A}}_i}({\bf{x}} - ({{\bf{\mu }}_i} + {\bf{\bar x}}))\} } \\
		= \sum\limits_{i = 1}^m {{w_i}\exp \{  - {{({\bf{x}} - {{{\bf{\bar \mu }}}_i})}^T}{{\bf{A}}_i}({\bf{x}} - {{{\bf{\bar \mu }}}_i})\} } 
	\end{array},
\end{equation}

\noindent where ${{{\bf{\bar \mu }}_i} = {{\bf{\mu }}_i} + {\bf{\bar x}}}$.
\emph{2) Image Scaling}

With the scaling factor ${k}$, the image scaling can be implemented by adjusting both the geometric center ${{{\bf{\mu }}_i}}$ and precision matrix ${{\bf{A}_i}}$ as follows:
\begin{equation}
	\begin{array}{c}
		EmFace(k{\bf{x}})\\
		= \sum\limits_{i = 1}^m {{w_i}\exp \{  - {{(k{\bf{x}} - {{\bf{\mu }}_i})}^T}{{\bf{A}}_i}(k{\bf{x}} - {{\bf{\mu }}_i})\} } \\
		= \sum\limits_{i = 1}^m {{w_i}\exp \{  - {{({\bf{x}} - \frac{1}{k}{{\bf{\mu }}_i})}^T}({k^2}{{\bf{A}}_i})({\bf{x}} - \frac{1}{k}{{\bf{\mu }}_i})\} } \\
		= \sum\limits_{i = 1}^m {{w_i}\exp \{  - {{({\bf{x}} - {{{\bf{\bar \mu }}}_i})}^T}{{{\bf{\bar A}}}_i}({\bf{x}} - {{{\bf{\bar \mu }}}_i})\} } 
	\end{array},
\end{equation}

\noindent where ${{{\bf{\bar \mu }}_i} = \frac{1}{k}{{\bf{\mu }}_i}}$, ${{{\bf{\bar A}}_i} = {k^2}{{\bf{A}}_i}}$.

\emph{3) Image Rotation}

With the rotation angle ${{\theta}}$, the image rotation around ${{\bf{\bar x}} = \left[ {\begin{array}{*{20}{c}}
			{{{\bar x}_1}}\\
			{{{\bar x}_2}}
	\end{array}} \right]}$  is implemented by adjusting both the geometric center ${{{\bf{\mu }}_i}}$ and precision matrix ${{\bf{A}_i}}$ as follows:
\begin{equation}
	\begin{array}{c}
		EmFace(({{\bf{F}}_r} \cdot ({\bf{x}} - {\bf{\bar x}})) + {\bf{\bar x}})\\
		= \sum\limits_{i = 1}^m {{w_i}\exp \{  - {{({{\bf{F}}_r} \cdot ({\bf{x}} - {\bf{\bar x}}) - {{\bf{\mu }}_i} + {\bf{\bar x}})}^T}} \\
		{\kern 1pt} {\kern 1pt} {\kern 1pt} {\kern 1pt} {\kern 1pt} {\kern 1pt} {\kern 1pt} {\kern 1pt} {\kern 1pt} {\kern 1pt} {\kern 1pt} {\kern 1pt} {\kern 1pt} {\kern 1pt} {\kern 1pt} {\kern 1pt} {\kern 1pt} {\kern 1pt} {\kern 1pt} {\kern 1pt} {\kern 1pt} {\kern 1pt} {\kern 1pt} {\kern 1pt} {\kern 1pt} {\kern 1pt} {\kern 1pt} {\kern 1pt} {\kern 1pt} {\kern 1pt} {\kern 1pt}  \cdot {{\bf{A}}_i} \cdot ({{\bf{F}}_r} \cdot ({\bf{x}} - {\bf{\bar x}}) - {{\bf{\mu }}_i} + {\bf{\bar x}})\} \\
		= \sum\limits_{i = 1}^m {{w_i}\exp \{  - {{({\bf{x}} - {{\bf{F}}_r}^{ - 1} \cdot ({{\bf{\mu }}_i} + {{\bf{F}}_r} \cdot {\bf{\bar x}} - {\bf{\bar x}}))}^T}} \\
		 {\kern 1pt} {\kern 1pt} {\kern 1pt} {\kern 1pt} {\kern 1pt} {\kern 1pt} {\kern 1pt} {\kern 1pt} {\kern 1pt} {\kern 1pt} {\kern 1pt} {\kern 1pt} {\kern 1pt} {\kern 1pt} {\kern 1pt} {\kern 1pt} {\kern 1pt}  \cdot {{\bf{F}}_r}^T{{\bf{A}}_i}{{\bf{F}}_r} \cdot ({\bf{x}} - {{\bf{F}}_r}^{ - 1} \cdot ({{\bf{\mu }}_i} + {{\bf{F}}_r} \cdot {\bf{\bar x}} - {\bf{\bar x}}))\} \\
		= \sum\limits_{i = 1}^m {{w_i}\exp \{  - {{({\bf{x}} - {{{\bf{\bar \mu }}}_i})}^T} \cdot {{{\bf{\bar A}}}_i} \cdot ({\bf{x}} - {{{\bf{\bar \mu }}}_i})\} } 
	\end{array},
\end{equation}

\noindent where ${{{\bf{F}}_r} = \left[ {\begin{array}{*{20}{c}}
			{\cos \theta }&{\sin \theta }\\
			{ - \sin \theta }&{\cos \theta }
	\end{array}} \right]}$ denotes the rotation matrix, and ${{{\bf{\bar \mu }}_i} = {{\bf{F}}_r}^{ - 1} \cdot ({{\bf{\mu }}_i} + {{\bf{F}}_r} \cdot {\bf{\bar x}} - {\bf{\bar x}})}$, ${{{\bf{\bar A}}_i} = {{\bf{F}}_r}^T{{\bf{A}}_i}{{\bf{F}}_r}}$.

\section{Experiments}\label{sec5}

Four types of experiments were performed to validate the proposed EmFace and EmNet. First, we analyzed the representation power of EmFace with different numbers of function elements by examining how well EmFace can reconstruct a face image from a learned latent embedding. Second, we compared the performance of our method to that of the baselines on several face image datasets. Third, we executed our EmNet on corrupted face images to achieve face image restoration and denoising, and compared the performance with that of several AE-based methods. Finally, we examined the capabilities of image transformations using EmFace. To further demonstrate the modeling results of EmFace, different samples were selected in each experiment to display quantitative results.

\textbf{Datasets}. Unless specified otherwise, the experiments were conducted on the LFW dataset \cite{huang2008labeled}. All 13,233 face images of 5,749 subjects in the LFW dataset, which were collected from the web under wild conditions, were used to validate the effectiveness of EmFace. In addition, we tested the performance of the proposed method on very challenging large-scale databases, including IARPA Janus Benchmark-B (IJB-B) and IJB-C \cite{whitelam2017iarpa}.

\textbf{Baselines}. Because there are no published reports on explicit models for face image modeling, we compared most of our results with those of two classical image code methods, namely, PCA \cite{turk1991eigenfaces} and SC \cite{Bao2016Dictionary}. We also compared the AE-based methods \cite{0Reducing, 2014Stacked, 2010Stacked}, which encode face images into hidden variables and then decode the hidden variables into face images.

\textbf{Experimental Setup}. In all the experiments, each face image was first converted to a gray image and then cropped to ${120 \times 120}$ pixels to remove background information. We trained EmNet on the VGG dataset \cite{parkhi2015deep}. In the SC method, the VGG dataset was used as a training set to learn the dictionary; face images were first sampled as ${9 \times 9}$ blocks, the size of the dictionary was set to 1,024, and the image was then reconstructed with sparse coefficients in the dictionary. In the PCA method, the VGG dataset was used as a training set to obtain the eigenvector matrix, and the principal components were then set to reconstruct the face. In the AE-based methods, we used the same ResNet50 architecture \cite{he2016deep} as our EmFace encoder and a ResNet50 symmetric architecture as the decoder, and the network was also trained on VGG.

\subsection{Experiment for Representation Power}
\begin{figure}[t]
	\centering
	\includegraphics[width=1.0\linewidth]{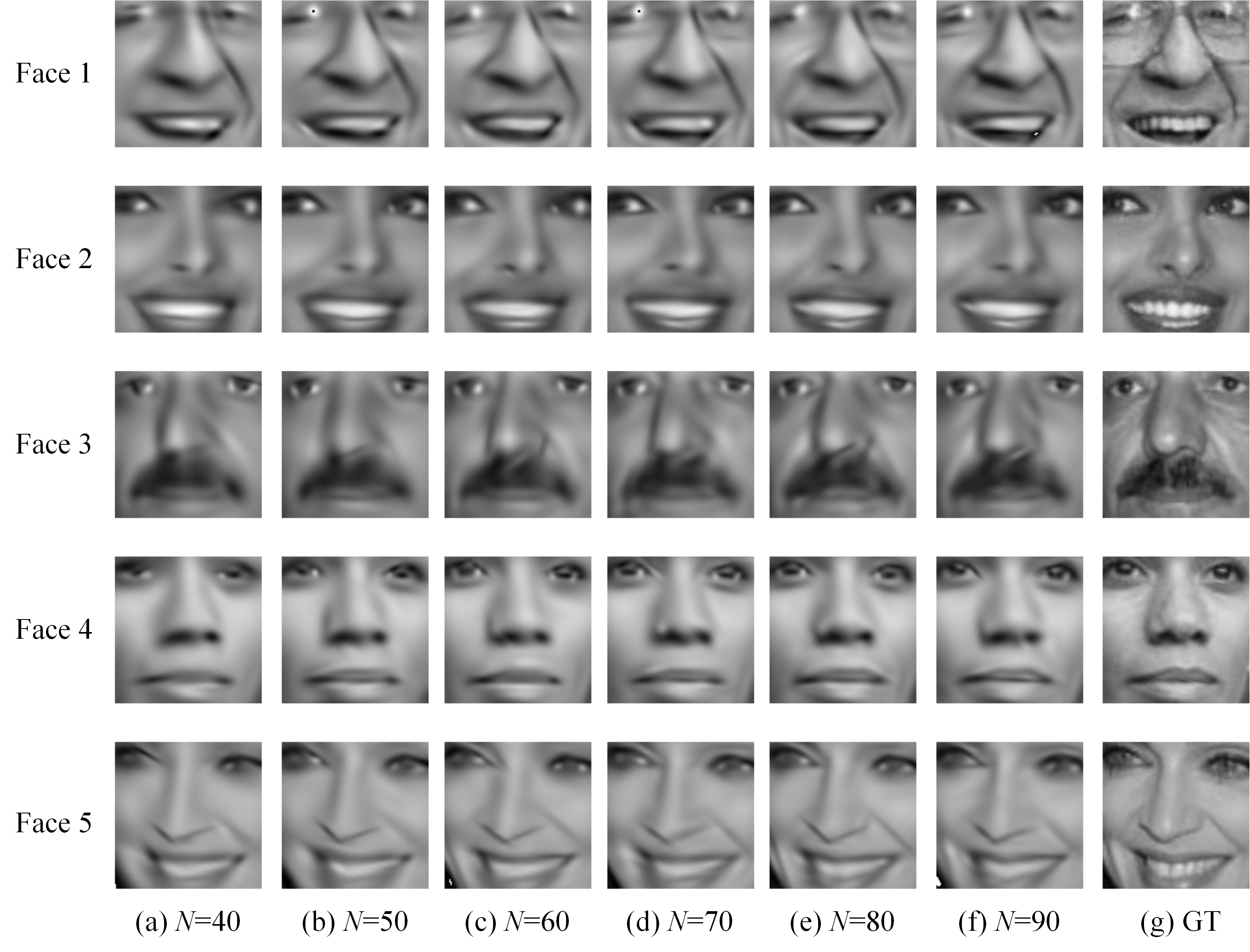}
	\caption{Qualitative results of face images by EmFace with 40, 50, 60, 70, 80 and 90 function elements.}\label{fig:the_nunber_of_function_elements_1}	
\end{figure}

\begin{figure}[t]
	\centering
	\includegraphics[width=1.0\linewidth]{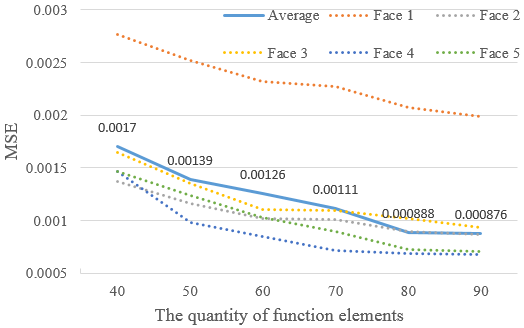}
	\caption{Quantitative results of face images by EmFace with 40, 50, 60, 70, 80 and 90 function elements.}\label{fig:the_nunber_of_function_elements_2}	
\end{figure}

In our first experiment, we investigated how well EmNet networks represent face images. We attempted to explore whether our network could learn a memory-efficient representation of a face image with a limited number of function elements. Through this experiment, we could determine the number of function elements ${N}$ in EmFace. We trained our neural network to embed each training sample in a ${6 \times N}$ dimensional latent space and reconstruct the face image from this embedding. Fig. \ref{fig:the_nunber_of_function_elements_1} shows the qualitative results for examples from the LFW dataset, with 40, 50, 60, 70, 80, and 90 function elements. Additionally, we measured the corresponding MSEs with respect to the ground truth image, as shown in Fig. \ref{fig:the_nunber_of_function_elements_2}. The following conclusions can be drawn from the results:
\begin{itemize}
	\item [1)]
	EmFace calculated by EmNet can achieve reasonable qualitative and quantitative results with a limited number of function elements.
	
	\item [2)]
	With an increase in the number of function elements, the size of ${\Theta}$ gradually increases, and the evaluation index MSE decreases. The qualitative results indicate that the images reconstructed by EmFace are similar to the original images, particularly in the detail location.
	
	\item [3)]
	3)	When the number of function elements reaches 80, the qualitative results are significantly improved. Beyond this number, a further increase in elements provides little improvement. Therefore, in the remaining experiments, the number of function elements ${N}$ was set to 80, and the corresponding ${\Theta}$ was 480-D.
\end{itemize}

\begin{table}[!htbp]
	\caption{ Comparison of the average mean square error (MSE) on the entire LFW database}
	\label{table:averageMSELFW}
	\centering
	\begin{tabular}{m{2.0cm}<{\centering} m{2.0cm}<{\centering}}
		\hline
		{ Methods} &average MSE\\
		\hline
		{PCA}&{\centering $0.001434$ }\\
		{SC}&{\centering $0.000931$ }\\
		{AE}&{\centering $0.001191$ }\\
		{DAE}&{\centering $0.001134$ }\\
		{\textbf{EmFace}}&{\centering $\textbf{0.000888}$ }\\
		\hline
	\end{tabular}
\end{table}

\begin{figure}[!htbp]
	\centering
	\includegraphics[width=0.7\linewidth]{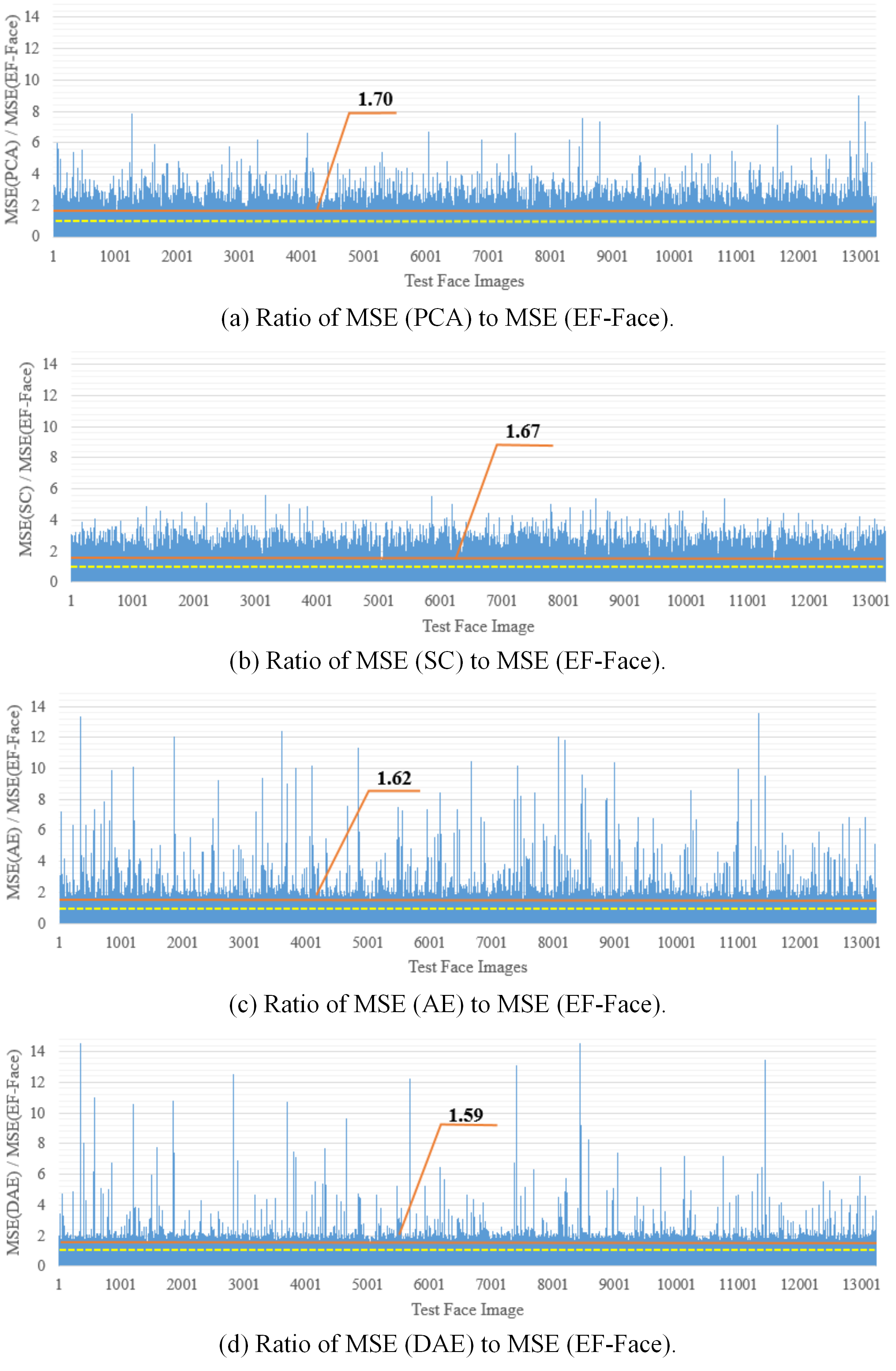}
	\caption{MSE ratios for all 13,233 face images in LFW. The orange line represents the mean of the ratios and the yellow dotted line signifies that the MSE ratio is equal to 1. Sub-figures respectively correspond to the ratio of MSE (PCA) to MSE (EmFace), the ratio of MSE (SC) to MSE (EmFace), the ratio of MSE (CAE) to MSE (EmFace), and the ratio of MSE (DAE) to MSE (EmFace).}
	\label{fig:fig_ratio_LFW}
\end{figure}
\subsection{Comparison with the baselines}

\begin{table*}[!htbp]	
	\caption{Representation results on some face image samples of LFW by EmFace, PCA, SC, and AE-based methods with the same parameter size (480).}	
	\label{table:Test_on_LFW}\small	
	\centering 
	\begin{tabular}{m{0.8cm}<{\centering}m{0.6cm}<{\centering}m{0.6cm}<{\centering}m{1.2cm}<{\centering}m{1.2cm}<{\centering}m{1.2cm}<{\centering}m{1.2cm}<{\centering}m{1.2cm}<{\centering}m{1.2cm}<{\centering}m{1.2cm}<{\centering}m{1.2cm}<{\centering}}
		\hline
		\multicolumn{3}{c}{\multirow{2}{*}{Methods}} & \multicolumn{8}{c}{Face images under the influence of different factors}\\
		\cline{4-11}
		&&&\multicolumn{2}{c}{normal}& \multicolumn{2}{c}{various postures}& \multicolumn{2}{c}{{\tabincell{c}{uneven\\ illumination}}}& \multicolumn{2}{c}{{\tabincell{c}{various \\expressions}}}\\
		\hline
		&&&&&&&&\\[-1.8ex]
		\multicolumn{3}{c}{original image}
		&\includegraphics[width=0.09\textwidth]{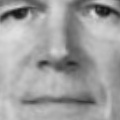}	
		&\includegraphics[width=0.09\textwidth]{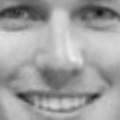}
		&\includegraphics[width=0.09\textwidth]{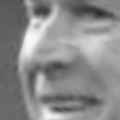}	
		&\includegraphics[width=0.09\textwidth]{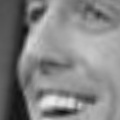} &\includegraphics[width=0.09\textwidth]{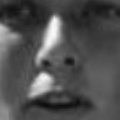}
		&\includegraphics[width=0.09\textwidth]{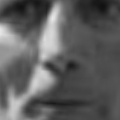}	
		&\includegraphics[width=0.09\textwidth]{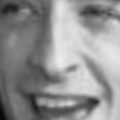} &\includegraphics[width=0.09\textwidth]{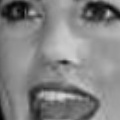}\\	
		\hline
		\multicolumn{2}{c}{\multirow{2}{*}[-14pt]{PCA}}&MSE&0.00159&0.00113&0.00216&0.00259&0.00248&0.00170&0.00157&0.00184\\
		&&Visual Effect
		&\includegraphics[width=0.09\textwidth]{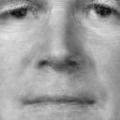}	
		&\includegraphics[width=0.09\textwidth]{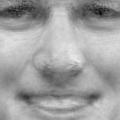}
		&\includegraphics[width=0.09\textwidth]{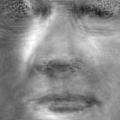}	
		&\includegraphics[width=0.09\textwidth]{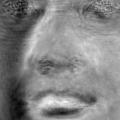}	
		&\includegraphics[width=0.09\textwidth]{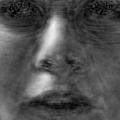}	
		&\includegraphics[width=0.09\textwidth]{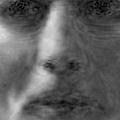}	
		&\includegraphics[width=0.09\textwidth]{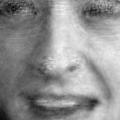}	
		&\includegraphics[width=0.09\textwidth]{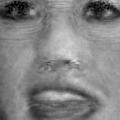}	\\		\hline
		\multicolumn{2}{c}{\multirow{2}{*}[-14pt]{SC}}&MSE&0.00113&0.00126&0.00110&0.000935&0.000567&0.000613&0.00118&0.00131\\
		&&Visual Effect
		&\includegraphics[width=0.09\textwidth]{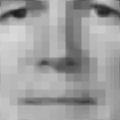}	
		&\includegraphics[width=0.09\textwidth]{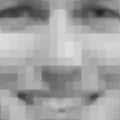}
		&\includegraphics[width=0.09\textwidth]{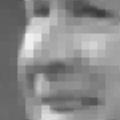}	
		&\includegraphics[width=0.09\textwidth]{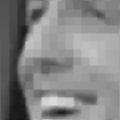}	
		&\includegraphics[width=0.09\textwidth]{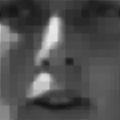}	
		&\includegraphics[width=0.09\textwidth]{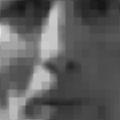}	
		&\includegraphics[width=0.09\textwidth]{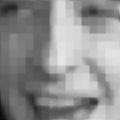}	
		&\includegraphics[width=0.09\textwidth]{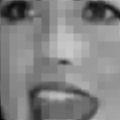}	\\
		\hline
		\multirow{4}{*}[-30pt]{\tabincell{c}{AE-based\\methods}}&\multirow{2}{*}[-7pt]{CAE}&MSE&0.0010&0.000702&0.001948&0.0024&0.0021&0.0014&0.0022&0.0056\\
		&&Visual Effect
		&\includegraphics[width=0.09\textwidth]{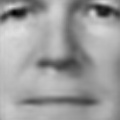}	
		&\includegraphics[width=0.09\textwidth]{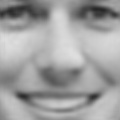}
		&\includegraphics[width=0.09\textwidth]{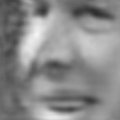}	
		&\includegraphics[width=0.09\textwidth]{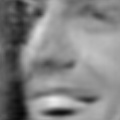}	
		&\includegraphics[width=0.09\textwidth]{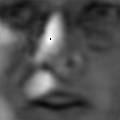}	
		&\includegraphics[width=0.09\textwidth]{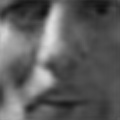}	
		&\includegraphics[width=0.09\textwidth]{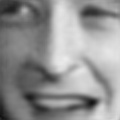}	
		&\includegraphics[width=0.09\textwidth]{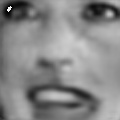}	\\
		&\multirow{2}{*}[-7pt]{DAE}&MSE&0.000973&0.000687&0.001963&0.0023&0.0018&0.000918&0.00104&0.0016\\
		&&Visual Effect
		&\includegraphics[width=0.09\textwidth]{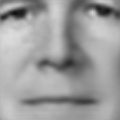}	
		&\includegraphics[width=0.09\textwidth]{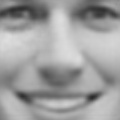}
		&\includegraphics[width=0.09\textwidth]{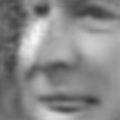}	
		&\includegraphics[width=0.09\textwidth]{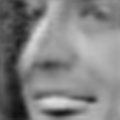}	
		&\includegraphics[width=0.09\textwidth]{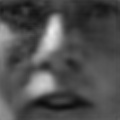}	
		&\includegraphics[width=0.09\textwidth]{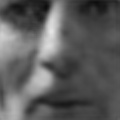}	
		&\includegraphics[width=0.09\textwidth]{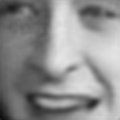}	
		&\includegraphics[width=0.09\textwidth]{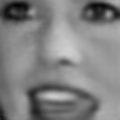}	\\		
		\hline
		\multicolumn{2}{c}{\multirow{2}{*}[-14pt]{\textbf{EmFace}}}&MSE&\textbf{0.000607}&\textbf{0.000543}&\textbf{0.000739}&\textbf{0.000858}&\textbf{0.000721}&\textbf{0.000526}&\textbf{0.000716}&\textbf{0.00126}\\
		&&Visual Effect
		&\includegraphics[width=0.09\textwidth]{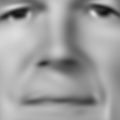}	
		&\includegraphics[width=0.09\textwidth]{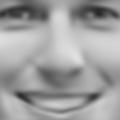}
		&\includegraphics[width=0.09\textwidth]{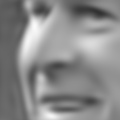}	
		&\includegraphics[width=0.09\textwidth]{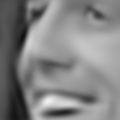}	
		&\includegraphics[width=0.09\textwidth]{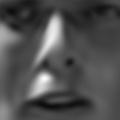}	
		&\includegraphics[width=0.09\textwidth]{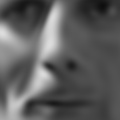}	
		&\includegraphics[width=0.09\textwidth]{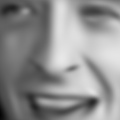}	
		&\includegraphics[width=0.09\textwidth]{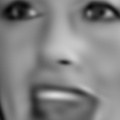}	\\	
		\hline
	\end{tabular}
\end{table*}

We evaluated the performance of EmFace, PCA, and several AE-based methods on various face image datasets. For the PCA method, the parameter size is the number of principal components, and for the AE-based methods, the parameter size is the length of the hidden variable of the encoder. For the sake of fairness, experiments were conducted with the same parameter size (480-D) to compare the representation effect between EmFace and the other methods.

\subsubsection{Test on LFW \cite{huang2008labeled} database}
The LFW \cite{huang2008labeled} database is the most commonly used classical face image database. First, we executed our EmFace and all other methods on the entire LFW database. Table \ref{table:averageMSELFW} lists the average MSE of our EmFace and those of the baselines. Our EmFace achieved a lower MSE than those of the PCA, SC, and all AE-based methods. PCA is a shallow linear dimension reduction method, and its MSE of face reconstruction using 480 -dimensional principal components is larger than that of other transformations. DAE, which is trained locally to denoise corrupted versions of inputs, can shape a more useful representation than the classical AE can, under the same network framework.

Subsequently, to evaluate the reconstruction effect of different methods, for every image, the MSE ratio of the comparison methods and EmFace (the former divided by the latter) was computed, as shown in Fig. \ref{fig:fig_ratio_LFW}.

As shown in Fig. \ref{fig:fig_ratio_LFW}, EmNet can faithfully represent the entire database. On average, the MSEs of PCA, SC, AE, and DAE are 1.70, 1.67, 1.62, and 1.59 times that of EmFace, respectively. There are 12,700 images (${95.97\%}$ of the total sample size) that exhibit more effective performance by EmFace than that of AE. Only 700 images were more accurately represented by DAE than EmFace, accounting for ${5.29\%}$ of the total sample size. For ${90.63\%}$ of the images, the MSE values of EmFace were less than those of PCA. Compared with PCA and AE-based methods, SC is more stable for the entire database, and ${83.00\%}$ of the images gain lower MSE values by EmFace than SC.

Subsequently, we selected eight face image samples that were more representative (with various postures, uneven illumination, and different expressions) to investigate the representation power of all the methods. Table \ref{table:Test_on_LFW} presents the qualitative and quantitative results.

According to Table \ref{table:Test_on_LFW}, EmNet enables us to represent the details of face images under the influence of various factors. Under the same parameter size, AE-based methods use a network symmetrical to ResNet50 to reconstruct the face image. To obtain an image of the same size, the upsampling operation is frequently applied; thus, the reconstructed images are fuzzier than those of EmNet. In particular, for images with large posture variations, uneven illumination, and exaggerated expressions, the AE-based methods are not able to represent them accurately. To satisfy the construction condition of the overcomplete dictionary, we refer to the general method of SC and perform image block coding; hence, the result of the SC method has an obvious block effect. As for PCA, the shallow linear transformation limits its representation, and most of the MSE values are greater than those of EmNet and AE-based methods under the same parameter size. In addition to the greater MSE values, the visual effect of the reconstructed images is ineffective, and there is an obvious noise artifact. 

\subsubsection{Test on challenging large-scale databases} 

\begin{table*}[!htbp]
	\caption{ Comparison of quantitative results on the IJB-B and IJB-C datasets}
	\label{table:test_results_ijb}
	\centering
	\begin{tabular}{cccc|ccc}
		\hline
		\multirow{3}{0.2\columnwidth}{\centering Methods} & \multicolumn{3}{c|}{IJB-B dataset} &\multicolumn{3}{c}{IJB-C dataset} \\
		& \multicolumn{3}{c|}{Number of images = ${76.8k}$} &\multicolumn{3}{c}{Number of images = ${148.8k}$} \\ 
		\cline{2-7}
		&average MSE$^a$ &{average ratio}$^b$&proportion$^c$&average MSE$^a$ &{average ratio}$^b$&proportion$^c$\\
		\hline
		{\centering PCA}&0.001657&2.24&96.34\%&0.001658&2.21&95.68\%\\
		{\centering SC}&0.001132&1.42&61.53\%&0.001050&1.35&57.96\%\\
		{\centering CAE}&0.001553&1.88&99.97\%&0.001558&1.89&99.94\%\\
		{\centering DAE}&0.001502&1.85&99.91\%&0.001511&1.83&99.89\%\\
		{\centering \textbf{EmFace}}&\textbf{0.000936}&1&-&\textbf{0.000953}&1&-\\
		\hline
	\end{tabular}
	\\[0.8ex]
	\footnotesize{$^a$ Average MSE represents the average value of MSEs on the entire database.}\\
	\footnotesize{$^b$ Average ratio denotes the average value of the ratio of MSE (baseline) to MSE (EmFace) on the entire database.}\\
	\footnotesize{$^c$ Proportion is the proportion of images with greater MSE (baseline) compared to MSE (EmFace) over the entire dataset.}\\	
\end{table*}

\begin{table*}[!htbp]	
	\caption{Representation results on some face image samples of IJB-B and IJB-C by EmFace, PCA, SC, and AE-based methods with the same parameter size (480).}	
	\label{table:test_results_ijb_1}\small	
	\centering 
	\begin{tabular}{m{0.9cm}<{\centering}m{0.6cm}<{\centering}m{0.6cm}<{\centering}m{1.2cm}<{\centering}m{1.2cm}<{\centering}m{1.2cm}<{\centering}m{1.2cm}<{\centering}|m{1.2cm}<{\centering}m{1.2cm}<{\centering}m{1.2cm}<{\centering}m{1.2cm}<{\centering}m{1.2cm}<{\centering}}
		\hline
		\multicolumn{3}{c}{\multirow{2}{*}{Methods}} & \multicolumn{8}{c}{Face images from different databases}\\
		\cline{4-11}
		&&&&&&&&\\[-1.7ex]
		&&& \multicolumn{4}{c|}{IJB-B}& \multicolumn{4}{c}{IJB-C}\\
		\hline
		&&&&&&&&\\[-1.8ex]
		\multicolumn{3}{c}{original image}	
		&\includegraphics[width=0.09\textwidth]{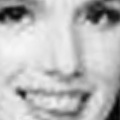}
		&\includegraphics[width=0.09\textwidth]{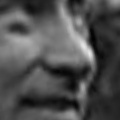}	
		&\includegraphics[width=0.09\textwidth]{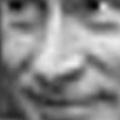} &\includegraphics[width=0.09\textwidth]{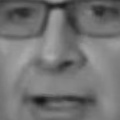}		
		&\includegraphics[width=0.09\textwidth]{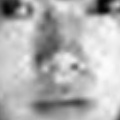}
		&\includegraphics[width=0.09\textwidth]{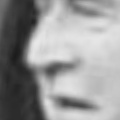}
		&\includegraphics[width=0.09\textwidth]{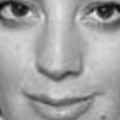} &\includegraphics[width=0.09\textwidth]{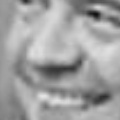}\\	
		\hline
		\multicolumn{2}{c}{\multirow{2}{*}[-14pt]{PCA}}&MSE&0.003427&0.002316&0.002493&0.004608&0.001030&0.002541&0.001178&0.001715\\
		&&Visual Effect	
		&\includegraphics[width=0.09\textwidth]{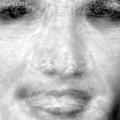}
		&\includegraphics[width=0.09\textwidth]{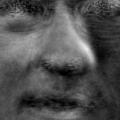}	
		&\includegraphics[width=0.09\textwidth]{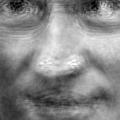}	
		&\includegraphics[width=0.09\textwidth]{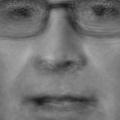}		
		&\includegraphics[width=0.09\textwidth]{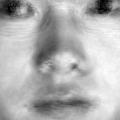}
		&\includegraphics[width=0.09\textwidth]{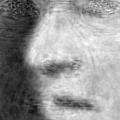}	
		&\includegraphics[width=0.09\textwidth]{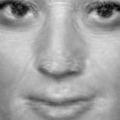}
		&\includegraphics[width=0.09\textwidth]{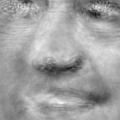}\\		
		\hline
		\multicolumn{2}{c}{\multirow{2}{*}[-14pt]{SC}}&MSE&0.001526&0.001317&0.001191&0.000682    &0.001412&0.000805&0.001444&0.001219\\
		&&Visual Effect	
		&\includegraphics[width=0.09\textwidth]{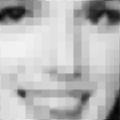}	
		&\includegraphics[width=0.09\textwidth]{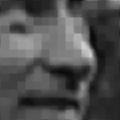}
		&\includegraphics[width=0.09\textwidth]{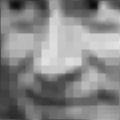}
		&\includegraphics[width=0.09\textwidth]{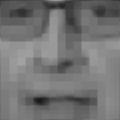}	
		&\includegraphics[width=0.09\textwidth]{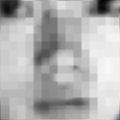}
		&\includegraphics[width=0.09\textwidth]{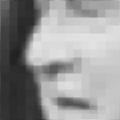}	
		&\includegraphics[width=0.09\textwidth]{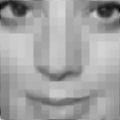}	
		&\includegraphics[width=0.09\textwidth]{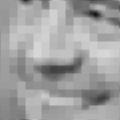}\\
		\hline
		\multirow{4}{*}[-30pt]{\tabincell{c}{AE-based\\methods}}&\multirow{2}{*}[-7pt]{CAE}&MSE&0.001704&0.002161&0.001689&0.006197  &0.001821&0.001542&0.001328&0.001783\\
		&&Visual Effect	
		
		&\includegraphics[width=0.09\textwidth]{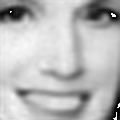}	
		&\includegraphics[width=0.09\textwidth]{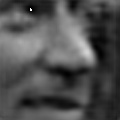}
		&\includegraphics[width=0.09\textwidth]{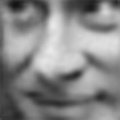}
		&\includegraphics[width=0.09\textwidth]{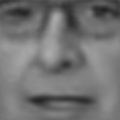}	
		&\includegraphics[width=0.09\textwidth]{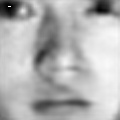}
		&\includegraphics[width=0.09\textwidth]{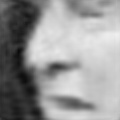}	
		&\includegraphics[width=0.09\textwidth]{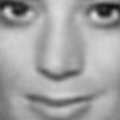}	
		&\includegraphics[width=0.09\textwidth]{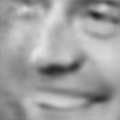}\\
		&\multirow{2}{*}[-7pt]{DAE}&MSE&0.001623&0.001968&0.001428&0.000557        &0.001863&0.001422&0.001248&0.001652\\
		&&Visual Effect
		
		&\includegraphics[width=0.09\textwidth]{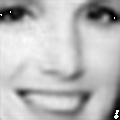}	
		&\includegraphics[width=0.09\textwidth]{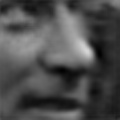}
		&\includegraphics[width=0.09\textwidth]{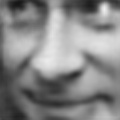}	
		&\includegraphics[width=0.09\textwidth]{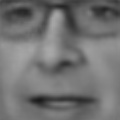}	
		&\includegraphics[width=0.09\textwidth]{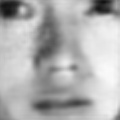}
		&\includegraphics[width=0.09\textwidth]{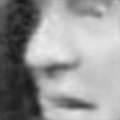}	
		&\includegraphics[width=0.09\textwidth]{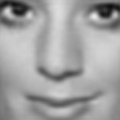}	
		&\includegraphics[width=0.09\textwidth]{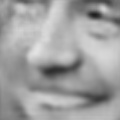}\\		
		\hline
		\multicolumn{2}{c}{\multirow{2}{*}[-14pt]{\textbf{EmFace}}}&MSE&\textbf{0.000917}&\textbf{0.001047}&\textbf{0.000812}&\textbf{0.000299}&\textbf{0.001044}&\textbf{0.000549}&\textbf{0.001004}&\textbf{0.000783}\\
		&&Visual Effect	
		&\includegraphics[width=0.09\textwidth]{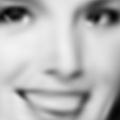}	
		&\includegraphics[width=0.09\textwidth]{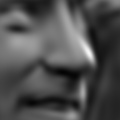}
		&\includegraphics[width=0.09\textwidth]{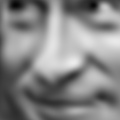}	
		&\includegraphics[width=0.09\textwidth]{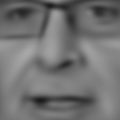}	
		&\includegraphics[width=0.09\textwidth]{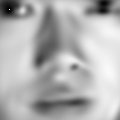}
		&\includegraphics[width=0.09\textwidth]{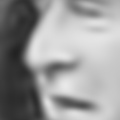}		
		&\includegraphics[width=0.09\textwidth]{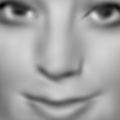}	
		&\includegraphics[width=0.09\textwidth]{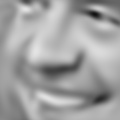}\\	
		\hline
	\end{tabular}
\end{table*} 


We tested the performance of the methods on very challenging large-scale databases, including the IJB-B and IJB-C datasets \cite{whitelam2017iarpa}. We did not use the IJB-A dataset \cite{klare2015pushing} because IJB-A has been superseded by IJB-B and forms a subset of IJB-B. The IJB-B dataset \cite{whitelam2017iarpa} is composed of face images with a large variety of facial postures and drastic changes in illumination under a completely unconstrained environment. It contains 1,845 subjects with 21.8k still images and 55k frames from 7,011 videos. The IJB-C dataset \cite{whitelam2017iarpa} is an extension of IJB-B, having 3,531 subjects with 31.3k still images and 117.5k frames from 11,779 videos. We ran our EmFace and the benchmark methods on all images and frames.

Table \ref{table:test_results_ijb} shows the quantitative results, and the three indexes are listed. The average MSE represents the average value of MSEs between the recovered face images and ground truths on the entire database. The average ratio denotes the average value of the ratio of MSE (baseline) to MSE (EmFace) for the entire database. The proportion is the proportion of images with the ratio of MSE (baseline) to MSE (EmFace) greater than one over the entire dataset.

We observed that our method achieved the lowest average MSE on both the IJB-B and IJB-C datasets. The average ratio of MSE (PCA), MSE (SC), MSE (CAE), and MSE (DAE) to MSE (EmFace) are 2.24, 1.42, 1.88, and 1.85 on IJB-B and 2.21, 1.35, 1.89, and 1.83 on IJB-C, respectively. Moreover, less than ${5.00\%}$ of the images exhibited greater MSE values by EmFace than PCA, CAE, and DAE on both the IJB-B and IJB-C datasets. Compared with SC, approximately ${60.00\%}$ of the images gained more effective performance by EmFace. In particular, compared to CAE and DAE, which have the same coder structure, EmFace achieved better representation results on almost all images. This shows that a continuous explicit model for face images learned by EmFace can achieve a more accurate representation than a neural network.

Table \ref{table:test_results_ijb_1} presents the visualization and representation loss for some selected face images. From this table, we can obtain a similar conclusion to that for the LFW dataset. In particular, the images in IJB-B and IJB-C were collected in a completely unconstrained environment. Therefore, we selected some low-quality images to examine the modeling effect. It is notable that the EmFace results are smoother than the original images in the entire face area in terms of visual effect.

\subsection{Face image restoration and denoising} 

In our third experiment, we conditioned EmNet on corrupted face images. The goal of this group of experiments was to evaluate how well EmNet can infer the correct face data from a corrupted input. Many studies have contributed to the development of image restoration and denoising, and the technology has made remarkable progress in the last few years \cite{jin2019flexible, quan2020self2self, 8481558, 8489894}. EmNet is not a method specially designed for image restoration and denoising; therefore, we only used DAE for comparison purposes. We trained EmNet and DAE on the VGG dataset and tested them on the LFW dataset with the same type of corrupted face images.

\begin{table}[!htbp]
	\caption{Quantitative results of face image restoration with two patch sizes}
	\label{table:image_restoration}
	\centering
	\begin{tabular}{m{1.5cm}<{\centering} m{2.5cm}<{\centering} m{2.5cm}<{\centering}}
		\hline
		\multirow{2}{*}{Methods} &\multicolumn{2}{c}{average MSE}\\
		\cline{2-3}
		&size of patch is ${40 \times 40}$ &size of patch is ${60 \times 60}$ \\
		\hline
		{DAE}&{\centering $0.001319$ }&{\centering $0.001663$ }\\
		\textbf{EmNet}&{\centering $\textbf{0.001029}$}&{\centering $\textbf{0.001427}$}\\
		\hline
	\end{tabular}
\end{table}

\begin{figure}[!htbp]
	\centering
	\includegraphics[width=1.0\linewidth]{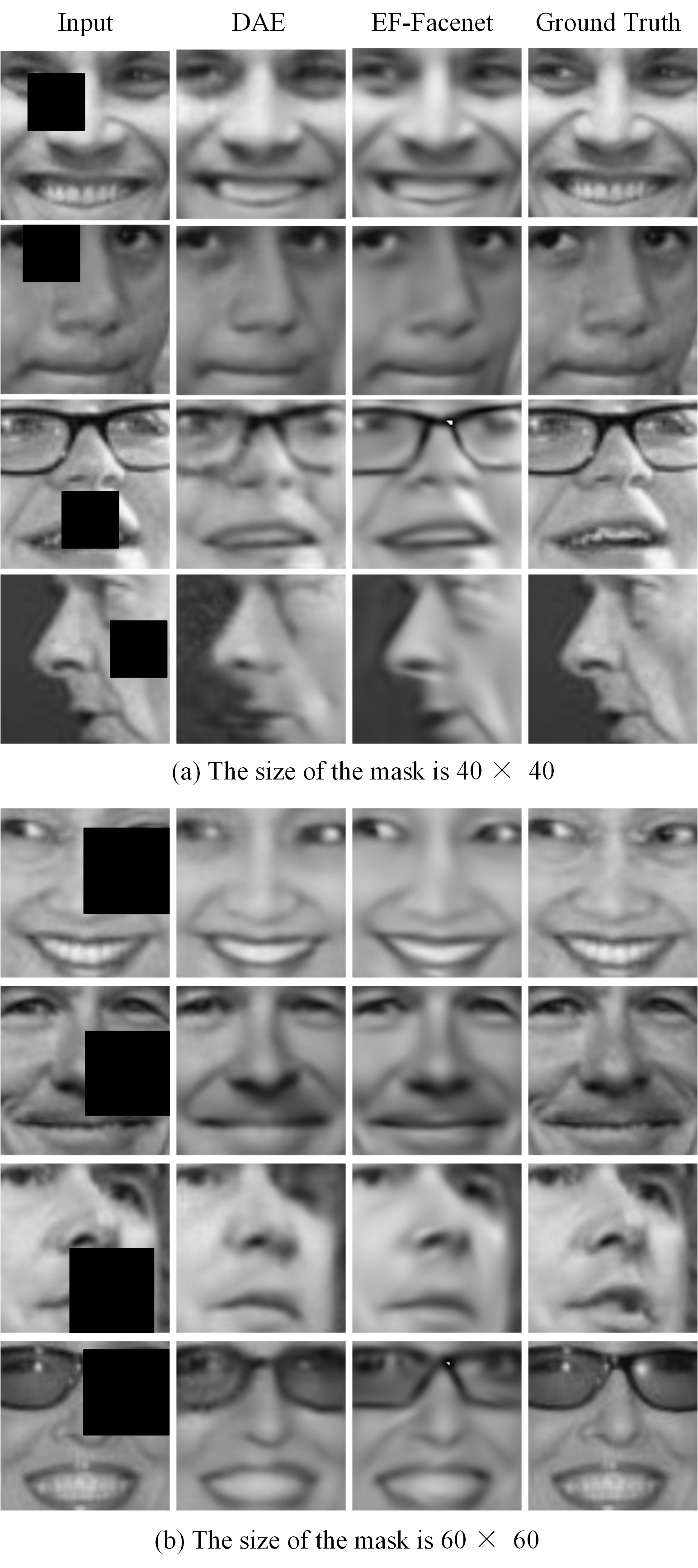}
	\caption{Qualitative results of face image restoration with two patch sizes}\label{fig:Restoration}	
\end{figure} 

\subsubsection{Face image restoration}

\begin{table}[!htbp]
	\caption{Quantitative results of AWGN denoising with two noise levels}
	\label{table:image_denoisy_AWGN}
	\centering
	\begin{tabular}{m{1.5cm}<{\centering} m{1.8cm}<{\centering} m{1.8cm}<{\centering}}
		\hline
		\multirow{2}{*}{Methods} &\multicolumn{2}{c}{average MSE}\\
		\cline{2-3}
		&${\sigma = 25}$ &${\sigma = 50}$ \\
		\hline
		{DAE}&{\centering $0.001232$ }&{\centering $0.001670$ }\\
		\textbf{EmNet}&{\centering $\textbf{0.000918}$}&{\centering $\textbf{0.001495}$}\\
		\hline
	\end{tabular}
\end{table}

\begin{figure}[!htbp]
	\centering
	\includegraphics[width=1.0\linewidth]{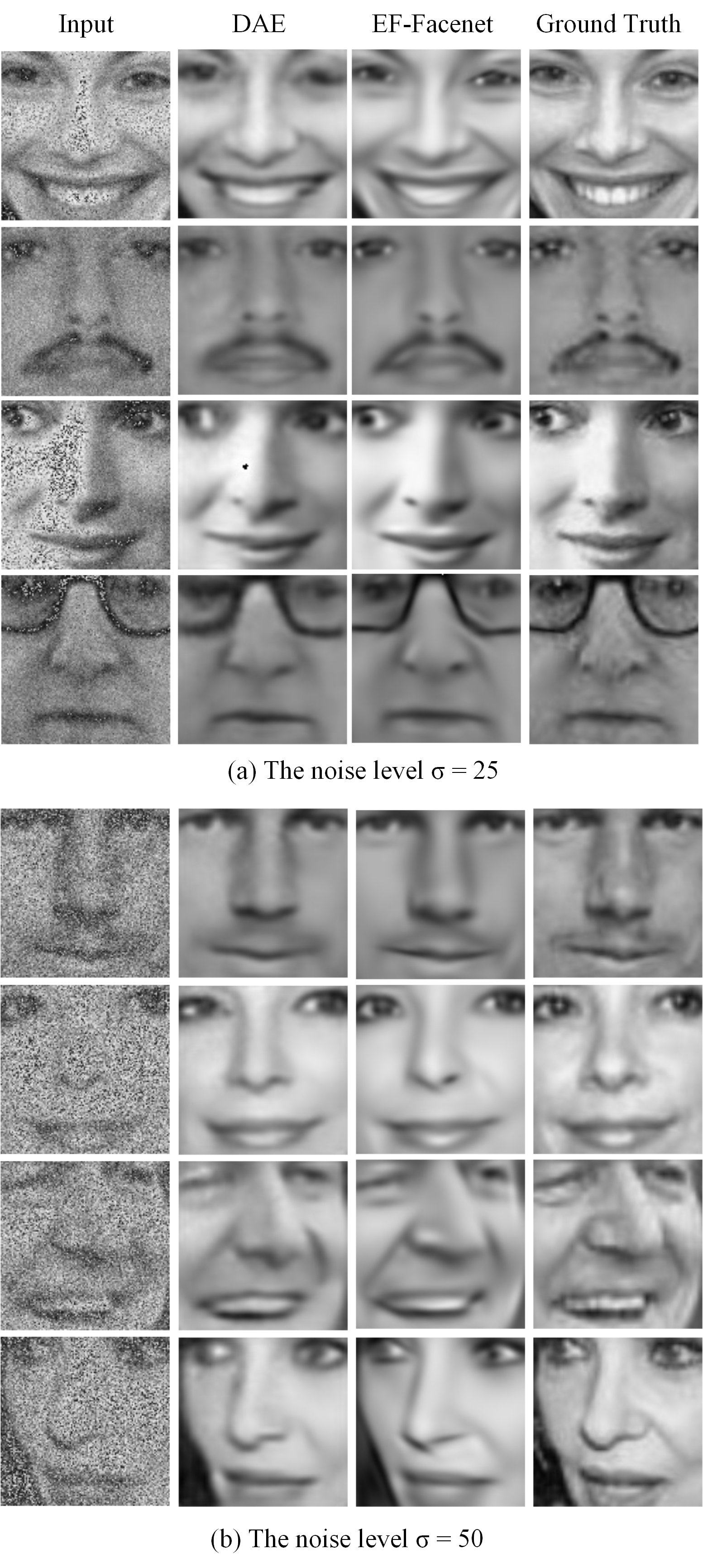}
	\caption{Qualitative results of AWGN denoising with two noise levels.}\label{fig:EFFacenet_denoisingAWGN}	
\end{figure}

In this experiment, we designed two patch sizes, ${40 \times 40}$ and ${60 \times 60}$. The mask was superimposed on a random position of the original image to generate corrupted face images for evaluation. EmNet attempts to restore the original image from corrupted face images. Table \ref{table:image_restoration} presents a quantitative comparison on the entire LFW database.

Compared with Table \ref{table:averageMSELFW}, the average MSEs of the two methods are larger than those of the direct modeling of the original images. Our method is considerably better than DAE for both the patch sizes. Fig. \ref{fig:Restoration} presents a visual comparison of several image samples. Both EmFace and DAE could restore the original image from occlusion; however, our method was better in terms of details.

\subsubsection{Face image denoising} 

The proposed method was evaluated on two face image denoising tasks: additive white Gaussian noise (AWGN) and salt-and-pepper noise removal (impulse noise) \cite{quan2020self2self}.

\textbf{AWGN denoising:} Current prevailing image denoising methods assume corrupted noise to be AWGN \cite{xu2018trilateral}. The degree of image damage was determined by the ${\sigma}$ of the Gaussian. In this experiment, corrupted images were generated by adding AWGN to the original images. We considered two noise levels, ${\sigma = 25, 50}$ \cite{jin2019flexible}.

Table \ref{table:image_denoisy_AWGN} presents the quantitative comparison, and Fig. \ref{fig:EFFacenet_denoisingAWGN} shows the visual comparison of several image samples. It can be seen from the results that the proposed method is superior to DAE in both quantitative and qualitative evaluations in removing two levels of AWGN.

\textbf{Salt-and-pepper noise:} Removing salt-and-pepper noise from images can be regarded as restoring randomly missing image pixels \cite{quan2020self2self}. The missing image pixels are random white or black spots. In this group of experiments, we randomly dropped the pixels of each image with the ratios of ${25\%}$ and ${50\%}$.

\begin{figure}[!htbp]
	\centering
	\includegraphics[width=1.0\linewidth]{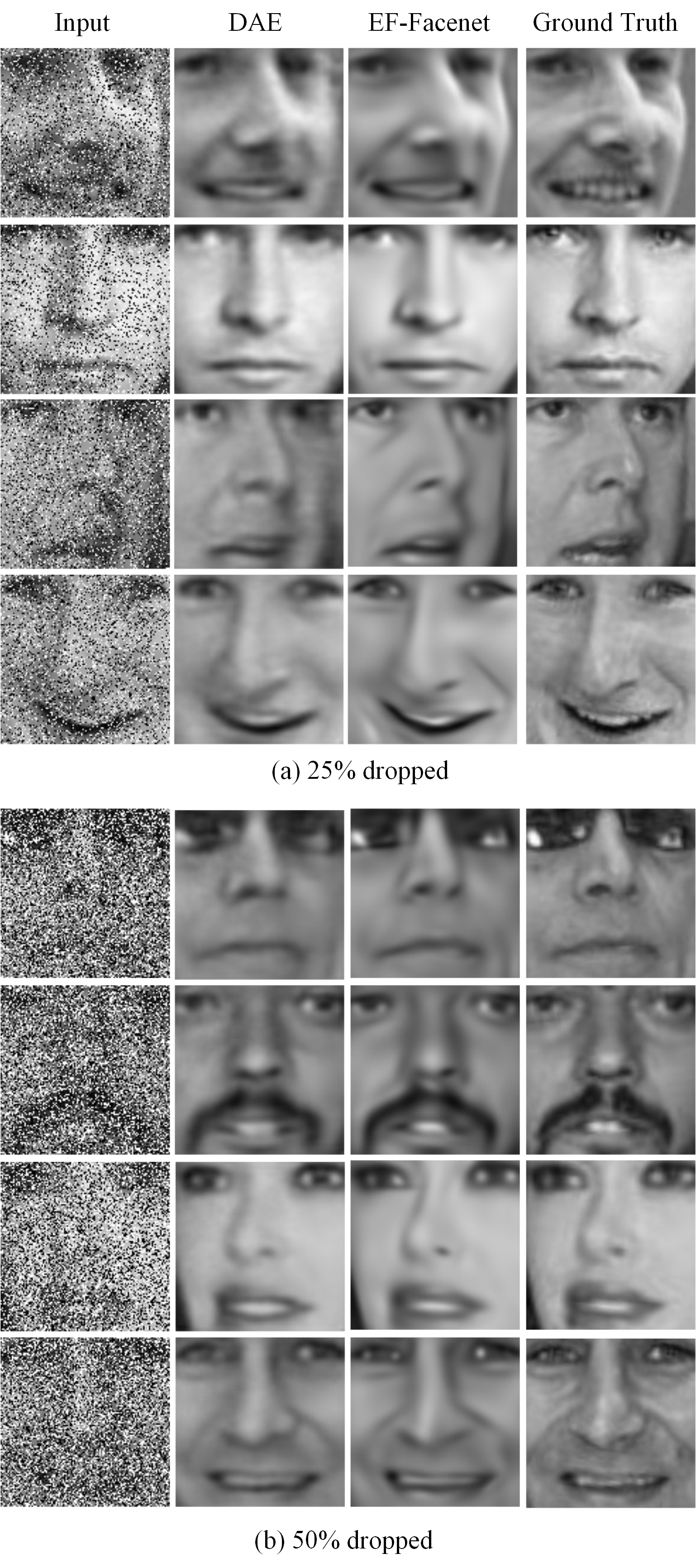}
	\caption{Qualitative results of salt-and-pepper denoising with two dropped levels.}\label{fig:EFFacenet_denoisingSaltandPepper}	
\end{figure}

We also evaluated the proposed method and DAE in terms of both quantitative and qualitative aspects. Table \ref{table:image_denoisy_SaltandPepper} lists the average MSE for the entire dataset. Our approach outperformed DAE at two noise levels, similar to face image restoration and AWGN denoising. Fig. \ref{fig:EFFacenet_denoisingSaltandPepper} demonstrates the visual comparison of ${25\%}$ and ${50\%}$ drops. The noise seriously affected the visual effect of the face image, particularly in the case of a ${50\%}$ drop. EmNet can infer the correct face data from the corrupted input well.

\begin{table}[!htbp]
	\caption{Quantitative results of salt-and-pepper denoising with two dropped levels}	
	\label{table:image_denoisy_SaltandPepper}
	\centering
	\begin{tabular}{m{1.5cm}<{\centering} m{1.8cm}<{\centering} m{1.8cm}<{\centering}}
		\hline
		\multirow{2}{*}{Methods} &\multicolumn{2}{c}{average MSE}\\
		\cline{2-3}
		&${25\%}$ dropped &${50\%}$ dropped\\
		\hline
		{DAE}&{\centering $0.001287$ }&{\centering $0.001497$ }\\
		\textbf{EmNet}&{\centering $\textbf{0.000887}$} &{\centering $\textbf{0.000925}$}\\
		\hline
	\end{tabular}
\end{table}

\subsection{Verification of Image Transformation}

\begin{figure}[!htbp]
	\centering
	\includegraphics[width=1.0\linewidth]{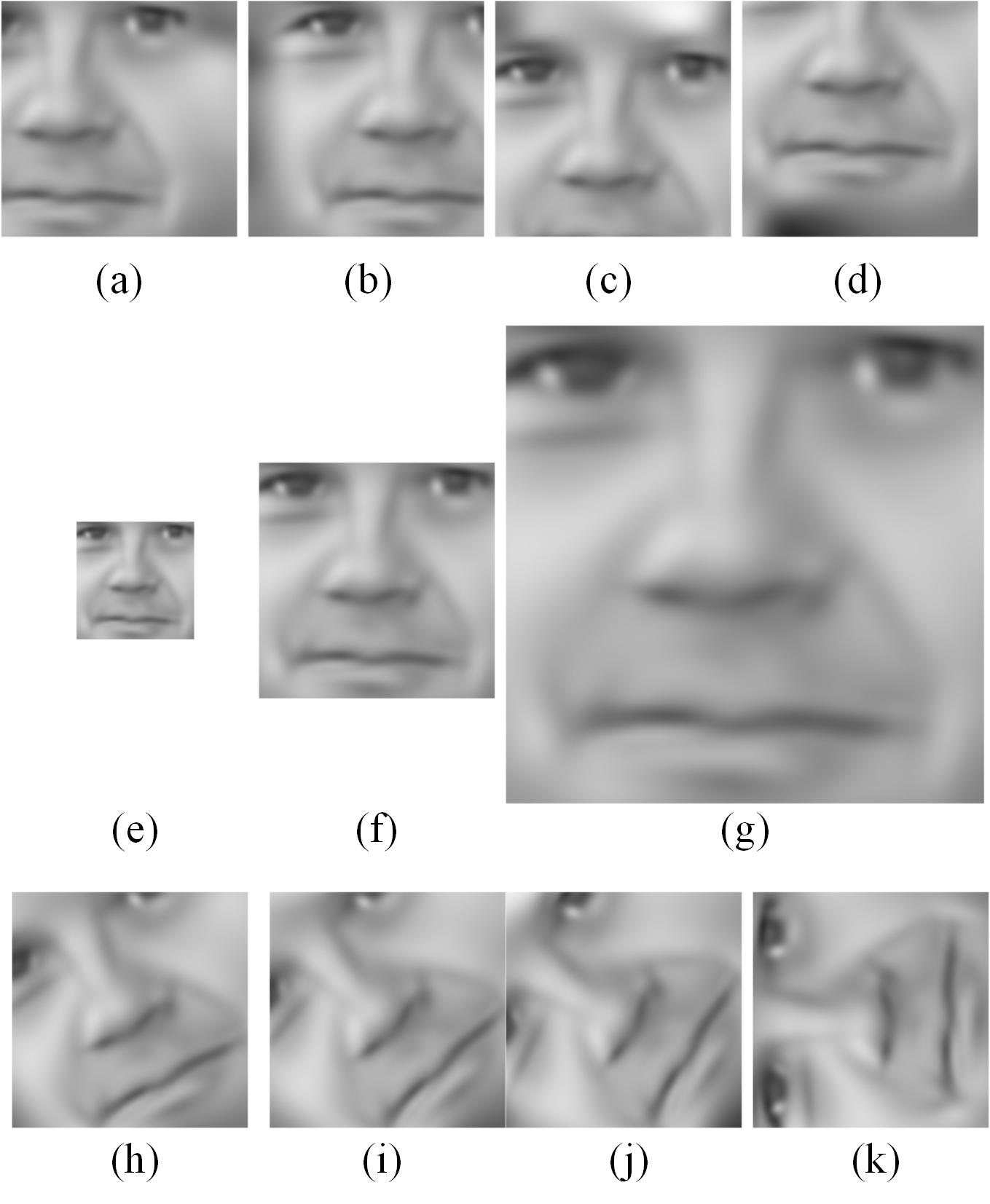}
	\caption{Verification of image transformation. The top row is the result of image translation processing through EmFace: (a) Move left. (b) Move right. (c) Move up. (d) Move down. The middle row is the result of image scaling processing through EmFace: (e) Scaling down. (f) EmFace. (g) Scaling up. The bottom row is the result of image rotation processing through EmFace: (h) Rotate ${\theta  = 30^\circ}$. (i) Rotate ${\theta  = 45^\circ}$. (j) Rotate ${\theta  = 60^\circ}$. (k) Rotate ${\theta  = 90^\circ}$.}
	\label{fig:Image_Transformation}
\end{figure}

Taking one face image in the LFW database as an example, based on the solved EmFace with 80 function elements, the image translation, scaling, and rotation can be realized using Eqs. (15), (16), and (17).

For image translation, the translational vector ${{\bf{\bar x}} = \left[ {\begin{array}{*{20}{c}}
			0\\
			{0.2}
	\end{array}} \right],\left[ {\begin{array}{*{20}{c}}
			0\\
			{ - 0.2}
	\end{array}} \right],\left[ {\begin{array}{*{20}{c}}
			{0.2}\\
			0
	\end{array}} \right],\left[ {\begin{array}{*{20}{c}}
			{ - 0.2}\\
			0
	\end{array}} \right]}$ represents that the translation direction, is left, right, up, and down, respectively. The image translation results are presented in Fig. \ref{fig:Image_Transformation} (a)–(d).

For image scaling, the scaling factor ${k = 2,\frac{1}{2}}$ denotes scaling down two times and scaling up two times, respectively. The image scaling results are presented in Fig. \ref{fig:Image_Transformation} (e) and (g).

For image rotation, the rotation angle ${\theta  = 30^\circ ,45^\circ ,60^\circ ,90^\circ }$ and the center of rotation ${{\bf{\bar x}} = \left[ {\begin{array}{*{20}{c}}{0.5}\\{0.5}\end{array}} \right]}$ imply that the image will rotate at different angles around the central point. The image rotation results are presented in Fig. \ref{fig:Image_Transformation} (h)–(k).

According to the results of the image transformation, the three operations of translation, scaling, and rotation can be achieved well by EmFace. Compared to the traditional methods of pixel operations, EmFace is a continuous transformation function of the gray value correlated with position. Based on EmFace, face images can be transformed through parameter adjustment without complicated interpolation transformations. Once the EmFace model of a face is constructed, the transformation becomes extremely simple. Moreover, as illustrated visually in the experimental results, EmFace has an additional smoothening effect.

\section{Conclusion}\label{sec7}

\textbf{Summary of research contributions:} In contrast to traditional methods based on handcrafted features and data-driven DNN learning methods, the explicit mathematical representation of human faces was investigated. We proposed EmFace as an explicit model of face representation, and a neural network EmNet was constructed as a parameter-solving network. Furthermore, EmNet supports applications such as face image restoration and denoising. Face image transformation can be realized mathematically using EmFace and simple parameter computations.

\textbf{Limitations and future work:} Although the representation results on experimental face images are encouraging, there are still several limitations that require further investigation. In this study, 2D face images were our modeling objects. As 3D face data could reflect more face pattern information, EmNet for 3D face modeling should be explored in the future. Farther, other bivariate elementary functions, such as the 2D Gabor function, trigonometric function, and exponential function, as function elements of EmFace, could be examined. In addition, based on our EmFace, more application tasks about face images such as face recognition, face detection, facial expression recognition, face generation will be further studied in the future. EmFace might not be the simplest mathematical model for face representation; however, it represents the first step toward this goal. 

\backmatter

\bmhead{Acknowledgments}

This work was supported by the National Science Foundation of China (Grant No. 61901436) and the Key Research Program of the Chinese Academy of Sciences (Grant No. XDPB22).

\bigskip

\bibliographystyle{sn-mathphys}
\bibliography{reference}


\end{document}